\pdfoutput=1

\documentclass[11pt]{article}

\usepackage[final]{ACL2023}

\usepackage{times}
\usepackage{latexsym}
\usepackage{amsmath}
\usepackage{amssymb}
\usepackage{booktabs}
\usepackage{graphicx}
\usepackage{array}
\usepackage{capt-of}
\usepackage[most]{tcolorbox}
\tcbset{
  rolloutlisting/.style={
    enhanced,
    breakable,
    colback=white,
    colframe=black,
    listing only,
    listing engine=listings,
    listing options={
      basicstyle=\ttfamily\footnotesize,
      breaklines=true,
      breakatwhitespace=false,
      columns=fullflexible,
      keepspaces=true,
      showstringspaces=false
    }
  }
}
\newcommand{\listingcaption}[2]{%
  \begin{center}
  \begin{minipage}{\linewidth}
  \captionof{figure}{#1}
  \label{#2}
  \end{minipage}
  \end{center}
}
\usepackage[T1]{fontenc}

\usepackage[utf8]{inputenc}

\usepackage{microtype}

\usepackage{inconsolata}

\title{S3C-LLM: Skill-Code Guided Agentic Language Models for Spectrum-to-Structure Elucidation}

\author{
   Xuanle Zhao\textsuperscript{\rm 1,2}, Xinyuan Cai\textsuperscript{\rm 1}\thanks{Corresponding Authors.}, Xiang Cheng\textsuperscript{\rm 1}, Bo Xu\textsuperscript{\rm 1,2}\footnotemark[\value{footnote}] \\
 \textsuperscript{\rm 1} The Key Laboratory of Cognition and Decision Intelligence for Complex Systems,\\ Institute of Automation, Chinese Academy of Sciences \\
 \textsuperscript{\rm 2} School of Artificial Intelligence, University of Chinese Academy of Sciences \\
\texttt{zhaoxuanle2022@ia.ac.cn}
}

\begin{document}
\maketitle
\begin{abstract}
Spectroscopic structure elucidation is central to molecular analysis, but recent Large Language Model (LLM)-based methods mostly formulate it as direct spectrum-to-SMILES generation.
Although this paradigm can leverage paired spectral data, it does not explicitly model the analytical workflow used by spectroscopists, such as diagnostic peak interpretation, fragment reasoning, formula constraints, and chemical consistency checking.
In this paper, we introduce S3C-LLM, a skill-guided and code-grounded agentic LLM for spectrum-to-structure elucidation.
Rather than directly predicting a molecule, S3C-LLM retrieves modality-specific spectroscopy skills, executes analysis code to instantiate these skills on the input spectra, and integrates the resulting peak-level evidence and formula constraints before generating SMILES.
Specifically, we contribute a self-evolving spectroscopy skill library, a thinking-augmented skill-code trajectory construction pipeline, and a two-stage training strategy that teaches Qwen3-4B through supervised fine-tuning (SFT) followed by our proposed step-level reinforcement learning (RL). Experiments on diverse benchmarks show that S3C-LLM consistently outperforms current general LLMs and spectrum-specific models across spectra, while using less than 1/10th of SpectraLLM's training corpus. 
\end{abstract}

\section{Introduction}

Automated molecular structure elucidation from spectra remains a central challenge in chemistry. Traditional systems rely on expert interpretation and spectral library search, making them labor-intensive to scale \citep{duhrkop2019sirius4}. Recent learning-based approaches reduce this reliance by directly modeling the mapping from spectra to molecular structures. For example, diffusion-based methods treat molecular generation as a conditional denoising process guided by spectral evidence \citep{bohde2025diffms,wang2025diffspectra}. More recently, LLM-based methods such as SpectraLLM \citep{su2026spectrallm} and SpecMol \citep{shen2025specmol} leverage the general capabilities of LLMs to enable multi-spectral molecular structure prediction.

Despite this progress, existing LLM-based methods largely follow a direct-answer paradigm, in which the model is trained to predict molecular strings directly from the input spectra. Although this paradigm can leverage existing paired spectrum-to-structure data, its end-to-end formulation does not explicitly model the process of spectroscopic analysis. In practice, spectroscopic structure elucidation is not merely a direct mapping from signals to molecular strings.  Human experts instead reason from diagnostic peaks to functional groups, fragments, and formula constraints before integrating the evidence into a chemically consistent structure. Recent spectroscopy agents such as LUMIR and IR-Agent follow a similar intuition by analyzing spectral evidence before prediction \citep{xie2025lumir,noh2025iragent}. However, their performance remains constrained by limited spectroscopy-specific chemical grounding.

To address these limitations, we introduce S3C-LLM, an agentic LLM that learns spectrum-to-structure prediction through a skill-guided, code-grounded reasoning process. We construct S3C-LLM through three steps: building a spectroscopy skill library via self-evolution, synthesizing agentic trajectories with executable analysis code, and training the model with supervised fine-tuning (SFT) followed by step-level reinforcement learning (RL).
First, we build a spectroscopy skill library that encodes spectroscopic prior knowledge. Each skill formalizes a reusable analysis pattern, such as linking diagnostic peak regions to functional groups, fragments, or substructure constraints, and is refined through prediction, verification, diagnosis, and revision. To make these skills operational, S3C-LLM generates and executes code guided by the retrieved skills to extract spectral evidence and perform quantitative checks. For example, in mass spectra, the code can compute mass differences and derive molecular-formula candidates under chemical-consistency constraints. We then use the evolved skills and code outputs to construct agentic reasoning trajectories, where each trajectory records how peak-level evidence and computational results support the final molecular prediction. To strengthen the reasoning process, we further generate thinking traces for each step by reverse-engineering the reasoning process from history content and reference action.
To train S3C-LLM, we first train the model to imitate skill-code agentic trajectories, and then use step-level RL to further optimize intermediate reasoning steps rather than only the final answer. This training strategy helps the model connect peak evidence, formula candidates, and chemical constraints to structure prediction. In conclusion, by learning the reasoning process behind structure elucidation, S3C-LLM remains data-efficient and achieves state-of-the-art performance across molecular structure elucidation metrics. With only 500K SFT data, less than 1/10th of SpectraLLM's reported paired-spectra corpus, S3C-LLM outperforms existing baselines. Our contributions are summarized as follows:
\begin{itemize}
      \item We introduce S3C-LLM, an agentic spectroscopy LLM that reformulates spectrum-to-structure prediction as a skill-guided and code-grounded reasoning process. It combines a self-evolving spectroscopy skill library with executable code analysis to construct interpretable agentic prediction trajectories.
      \item We propose a two-stage training strategy that first uses SFT to teach the model to imitate skill-code reasoning trajectories, and then applies step-level RL to improve intermediate reasoning and tool-use ability.
      \item Experiments on multiple spectral benchmarks show that S3C-LLM achieves state-of-the-art performance, outperforming task-specific backbones and requiring substantially fewer training samples than SpectraLLM.
\end{itemize}

\section{Related Work}

\subsection{Spectrum Elucidation Methods}

Conventionally, spectrum-to-structure elucidation is formulated as computer-assisted structure elucidation and chemically constrained candidate search. For example, in NMR, CASE systems use chemical shifts and two-dimensional correlations to constrain possible structures \citep{jaspars1999computer}, while spectral databases provide reference chemical-shift evidence for search and prediction \citep{steinbeck2003nmrshiftdb}. In MS/MS, CSI:FingerID \citep{duhrkop2015csifingerid} and SIRIUS \citep{duhrkop2019sirius4} rank candidates with molecular fingerprints and formula hypotheses. MetFrag \citep{ruttkies2016metfrag} and MS-FINDER \citep{tsugawa2016msfinder} use fragmentation rules for candidate scoring, and CFM-ID \citep{allen2015cfmid} models competitive fragmentation for metabolite identification. These methods provide strong chemical constraints, but usually rely on hand-designed rules, candidate libraries, or spectrum-specific search procedures.

Learning-based methods instead learn spectrum-to-structure relations from data. For NMR, substructure and multitask models support candidate ranking or direct structure elucidation \citep{huang2021nmrframework,hu2024accurate,yang2026nmrtrans}. For MS/MS, NEIMS and MassFormer learn structure-to-spectrum prediction \citep{wei2019neims,young2024massformer}, while DarkNPS, MSNovelist, and Spec2Mol generate molecular structures from spectra \citep{skinnider2021darknps,stravs2022msnovelist,litsa2023spec2mol}. MIST and DiffMS further introduce formula-aware transformers, attention-based generation, and diffusion generation for elucidating spectral structures \citep{goldman2023mist,bohde2025diffms}. More recently, multi-spectral benchmarks have broadened the evaluation beyond isolated spectra: the Multimodal Spectroscopic Dataset provides paired spectra across multiple modalities, and MassSpecGym focuses on molecule identification from experimental MS/MS spectra \citep{alberts2024unraveling,bushuiev2024massspecgym}. Built on this trend, SpectraLLM studies whether general-purpose LLMs can translate multi-spectral textual inputs into molecular structures, while SpecMol develops a spectroscopy-grounded foundation model for multi-task molecular learning \citep{su2026spectrallm,shen2025specmol}.

\subsection{Scientific LLM Agents}

Scientific LLM agents make reasoning processes more explicit by coupling LLM with external actions and tools. For example, ReAct interleaves reasoning with environment actions \citep{yao2023react}, and Toolformer learns when to invoke tools from self-supervised API annotations \citep{schick2023toolformer}. Beyond generic tool selection, program-aided language models delegate symbolic computation to executable code \citep{gao2023pal}. More recent skill-agent systems further move from one-shot tool calls toward persistent and validated skill libraries. For instance, SkillFoundry mines executable skills from heterogeneous scientific resources and CoEvoSkills constructs self-evolving skill packages with verifier feedback \citep{shen2026skillfoundry,zhang2026coevoskills}. These works motivate decomposing complex tasks into reusable actions.

Chemistry agents show similar benefits from domain operations and validation tools. ChemCrow integrates expert-designed chemistry tools for molecular search and synthesis-related operations \citep{bran2024chemcrow}, and Coscientist combines search, code execution, documentation reading, and laboratory automation \citep{boiko2023coscientist, zhao2026tinychemvl, zhao2026chemvlr}.

\section{Method}

\subsection{Overview}

Our S3C-LLM reformulates spectrum-to-structure prediction as a skill-guided, code-grounded reasoning process rather than a direct spectrum-to-SMILES mapping. Following the data setting of SpectraLLM \citep{su2026spectrallm}, we use samples from the Multimodal Spectroscopic Dataset \citep{alberts2024unraveling}, QM9S \citep{zou2023deep}, and MassSpecGym \citep{bushuiev2024massspecgym}, and convert raw spectra into compact peak-level representations. Given spectra from one or more modalities, S3C-LLM retrieves spectroscopy skills, executes code for spectrum-specific computations, and integrates peak-level evidence and formula constraints to predict the final molecular structure.

The whole framework follows three stages. We first build a spectroscopy skill library via self-evolution with a strong external LLM agent, then synthesize agentic trajectories with executable analysis code, and finally train the Qwen3-4B backbone with SFT followed by step-level RL. Let $x$ denote the processed spectral input and $y$ the target molecular structure. We represent each intermediate analysis trajectory as $\tau=\langle \mathcal{S}, c, o\rangle$, where $\mathcal{S}$ is the retrieved skill content, $c$ is the generated analysis code, and $o$ is the execution output. Thus, instead of only learning $p_{\theta}(y \mid x)$, S3C-LLM learns $p_{\theta}(\tau, y \mid x)$, so that the final prediction is conditioned on explicit skill-code evidence.

\subsection{Skill Library Construction}

Inspired by how spectroscopists analyze diagnostic peaks to infer functional groups, fragments, and molecular-formula candidates, we formalize this analysis knowledge into a spectroscopy skill library. Since paired spectral datasets rarely annotate intermediate analysis steps, we construct the library through self-evolution: an external LLM agent repeatedly applies a skill draft to spectra, checks whether the resulting interpretation is chemically consistent, and revises the skill when spectral rules are missing or misleading.

Concretely, we define a modality-specific skill template, write an initial draft for each modality, and use Claude Opus 4.6 with Claude Code \citep{anthropic2026claudeopus46,anthropic2026claudecode} as the external self-evolution agent. For each modality, the agent runs 20 evolution iterations with 50 samples per iteration, following a prediction-verification-diagnosis-revision loop. In each iteration, the agent applies the current skill to spectra, predicts the SMILES, checks whether the implied functional groups, fragments, and formula cues match the reference structure, and revises missing or misleading spectral rules when interpretation errors occur. The process yields eight modality-specific Markdown skills, each mapping spectral signals to chemical hypotheses such as functional groups, fragments, molecular-formula candidates, and substructure constraints. For example, the IR skill maps diagnostic and fingerprint frequency regions to functional-group hypotheses through single-peak and multi-peak rules, while the MS/MS skill maps precursor masses, neutral losses, and diagnostic fragment ions to molecular-formula candidates and fragment hypotheses.

\begin{figure*}[t]
\centering
\includegraphics[width=0.98\linewidth]{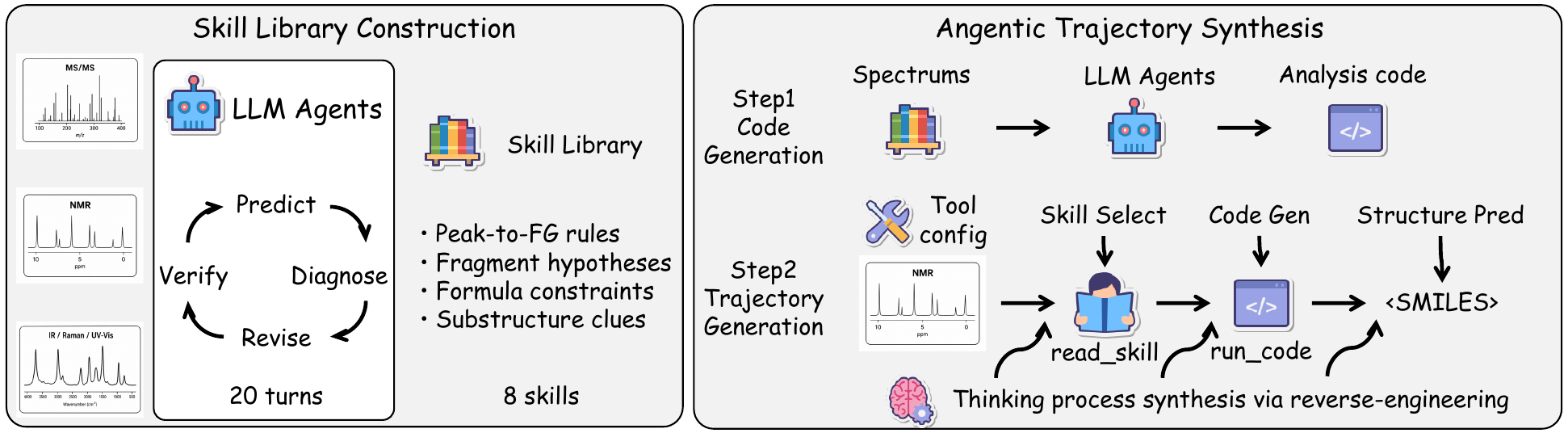}
\vspace{-5pt}
\caption{Overview of S3C-LLM data construction. We first construct a spectroscopy skill library via self-evolution, where an LLM agent applies, verifies, diagnoses, and revises modality-specific skills iteratively. Using the optimized skills, we generate executable analysis code, wrap skill retrieval and code execution into tool-use trajectories, and synthesize thinking traces to provide skill-code supervision for training.}
\vspace{-5pt}
\label{fig:data-construction}
\end{figure*}

\subsection{Agentic Data Construction}

Given the optimized skill library, we construct skill-code trajectories that turn each processed spectrum into a learnable analysis process for training S3C-LLM. We first use the Claude Code agent to generate executable analysis code for each spectrum, guided by the corresponding modality-specific skill. The generated code is designed to extract spectral evidence and perform quantitative checks. For instance, for MS/MS spectra, it computes mass differences, enumerates molecular-formula candidates, and compares fragments. For IR, Raman, UV-Vis, and NMR spectra, it parses peak or shift lists, matches diagnostic regions, and summarizes supported functional groups.

We then wrap skill retrieval and code execution into a tool-use trajectory. As shown in Table~\ref{tab:tool-config}, we define two tools for this process. The \texttt{read\_skill} tool retrieves one of eight modality-specific Markdown skill files, covering IR, Raman, UV-Vis, \textsuperscript{13}C NMR, \textsuperscript{1}H NMR, HSQC, simulated NMR, and MS/MS. The \texttt{run\_code} tool receives self-contained Python analysis code and returns the execution output as structured evidence.

For single-modality inputs, each trajectory calls \texttt{read\_skill} once to obtain the corresponding skill content $\mathcal{S}$, followed by one \texttt{run\_code} step. For inputs containing multiple spectral modalities, the trajectory simultaneously begins with multiple \texttt{read\_skill} calls, one for each modality, and still uses a single \texttt{run\_code} step to aggregate the returned skills and jointly analyze cross-modal evidence. We denote the generated code and its output as $c$ and $o$, and the model conditions on these skill-code observations to predict the final SMILES. Following this strategy, we construct 500k trajectories in the instruct mode.

\begin{table}[t]
\centering
\small
\setlength{\tabcolsep}{3pt}
\resizebox{\linewidth}{!}{
\begin{tabular}{>{\raggedright\arraybackslash}p{0.20\linewidth}>{\raggedright\arraybackslash}p{0.46\linewidth}>{\raggedright\arraybackslash}p{0.28\linewidth}}
\toprule
Tool & Arguments & Tool response \\
\midrule
\texttt{read\_skill} &
\texttt{name}: one of \texttt{\{msms, simnmr, ir, raman, uv, c\_nmr, h\_nmr, hsqc\}}. &
Markdown skill content $\mathcal{S}$. \\
\midrule
\texttt{run\_code} &
\texttt{code}: self-contained Python for evidence extraction and checks. &
Output $o$ with evidence and candidate constraints. \\
\bottomrule
\end{tabular}
}
\vspace{-5pt}
\caption{Tool interface, including the corresponding arguments and expected responses, used to construct S3C-LLM skill-code trajectories.}
\label{tab:tool-config}
\vspace{-5pt}
\end{table}

The resulting tool trajectories specify actions and observations, but the final prediction step can still lack explicit reasoning. Deducing the SMILES structure from spectra and code outputs remains highly uncertain. To strengthen this step, we synthesize thinking traces for sampled trajectories across different spectral modalities. Given the previous conversation history and the reference next action, we use GPT-5.4~\citep{openai2026gpt54} to reverse-engineer the thinking that should lead to that action, including intermediate tool calls and the final SMILES prediction. In particular, for the final step, the generated thinking is required to analyze the input spectrum, interpret the retrieved skill content, summarize the code output, and derive the SMILES step by step. We then use the same model to re-predict the SMILES from the generated thinking without access to the reference actions, and filter out trajectories whose prediction does not match the reference answer. This yields about 100K trajectories with explicit thinking traces. For the remaining 400K instruction trajectories, we keep empty \texttt{<think></think>} blocks to maintain a unified training schema, while applying explicit thinking supervision only to the filtered trajectories. Figure~\ref{fig:data-construction} summarizes the full data construction pipeline, including skill evolution, analysis-code generation, tool-use trajectory synthesis, and thinking-trace augmentation.

\subsection{Moldel Training}

\subsubsection{Supervised Fine-tuing}

We first perform SFT on the model using our constructed 500K agentic trajectories. All tool interactions are encapsulated within \texttt{<tool\_call>} blocks, including \texttt{read\_skill} for retrieving spectroscopy skills and \texttt{run\_code} for executing analysis code. To enhance the thinking process, we mask the training loss of empty \texttt{<think>} blocks in the instruct-mode trajectories, so that the model learns tool usage without being forced to imitate empty reasoning steps. Let $z=(\hat{\tau},y)$ denote the target sequence formed by the agentic trajectory and final SMILES. We optimize the masked token-level objective:
\begin{equation}
    \mathcal{L}_{\mathrm{SFT}}
    = - \sum_{(x,\hat{\tau},y)\in\mathcal{D}}
    \sum_{j=1}^{|z|}
    m_j \log p_{\theta}(z_j \mid x, z_{<j}),
\end{equation}
where $m_j=0$ for ignored empty-thinking tokens and $m_j=1$ otherwise.

\subsubsection{Step-Level RL}

Although SFT teaches S3C-LLM to imitate the constructed skill-code trajectories, pure imitation only matches reference actions and does not explicitly reward the quality of intermediate spectral reasoning. We therefore further refine the policy with Group Relative Policy Optimization (GRPO) \citep{shao2024deepseekmath} on 20K RL trajectories, where multiple rollouts are sampled for each spectral input and optimized with step-level verifiable rewards derived from the agentic analysis process. To construct these RL trajectories, we first run four preliminary rollouts for each candidate prompt and discard prompts whose four rollouts produce identical outcomes, filtering out cases with little group-relative training signal.

However, optimizing GRPO with only trajectory-level rewards is poorly matched to tool-augmented spectral reasoning. In online rollouts, failures are heterogeneous: about 10\% stem from tool-use errors, while most occur after valid tool execution and reflect incorrect final SMILES prediction. A trajectory-level reward entangles tool-use quality with final-structure correctness, causing correct intermediate actions to be penalized when the final prediction is wrong. We therefore propose step-level RL, which assigns credit at the granularity of assistant action steps and provides more localized supervision for spectral reasoning.

We decompose each rollout into assistant action steps, including skill retrieval, code execution, and final SMILES prediction. Each step receives a stage-specific verifiable reward, which is converted into a step advantage through reward-to-go estimation and same-step group normalization. This allows downstream structure-prediction quality to provide feedback to earlier skill-retrieval and code-execution actions, while avoiding a single coarse-grained reward for the whole trajectory.

Tool responses are appended as observations for subsequent assistant actions. A skill-retrieval step receives reward when the selected skill matches the required spectrum modality, and a code-execution step receives reward when the generated Python code executes without a tool error. The final prediction step is scored by Morgan fingerprint Tanimoto similarity, while malformed tool calls, missing required skills, failed execution, and invalid SMILES receive zero reward.

For the $i$-th rollout in a group of $M$ rollouts sampled from the old policy $\pi_{\theta_{\mathrm{old}}}$, let $a_{i,t}$ denote the assistant-token span of the $t$-th step and let $r_{i,t}$ denote its stage-specific reward. We compute the reward-to-go for each step as:
\begin{equation}
G_{i,t}=\sum_{k=t}^{T_i}\gamma^{k-t}r_{i,k}.
\end{equation}
The reward-to-go is then normalized among rollouts from the same prompt group at the same step index:
\begin{equation}
A_{i,t}=\frac{G_{i,t}-\mu_{g,t}}{\sigma_{g,t}+\epsilon_{\mathrm{norm}}},
\end{equation}
where $\mu_{g,t}$ and $\sigma_{g,t}$ are computed over rollouts from prompt group $g$ that contain step $t$. This preserves the group-relative comparison in GRPO, but compares actions that play similar roles in the agentic trajectory. We apply the return-based advantage $A_{i,t}$ to the assistant-token span $a_{i,t}$ that generates the $t$-th action and optimize the corresponding tokens with a clipped policy objective:
\begin{equation}
\begin{aligned}
\mathcal{J}_{\mathrm{RL}}(\theta)
=&\mathbb{E}\Bigg[
\frac{1}{M}\sum_{i=1}^{M}\sum_{t=1}^{T_i}
\frac{1}{|a_{i,t}|}\sum_{j\in a_{i,t}}\ell_{i,j}
\Bigg],
\\
\ell_{i,j}
=&\min\left(
\rho_{i,j}(\theta)A_{i,t},
\right.
\\
&\quad\left.
\operatorname{clip}\left(\rho_{i,j}(\theta),
1-\epsilon_{\mathrm{clip}},
1+\epsilon_{\mathrm{clip}}\right)A_{i,t}
\right).
\end{aligned}
\end{equation}
where the expectation is over input $x$ and rollouts sampled from the old policy:
\begin{equation}
\begin{aligned}
\{\hat{\tau}_i\}_{i=1}^{M}
&\sim \pi_{\theta_{\mathrm{old}}}(\cdot\mid x),\\
\rho_{i,j}(\theta)
&= \frac{\pi_{\theta}(z_{i,j}\mid h_{i,j})}
{\pi_{\theta_{\mathrm{old}}}(z_{i,j}\mid h_{i,j})},\\
h_{i,j} &= (x,z_{i,<j}).
\end{aligned}
\end{equation}
We do not use an explicit KL penalty in this setting. Since $A_{i,t}$ is computed from the reward-to-go, this objective allows downstream structure-prediction rewards to propagate to earlier skill-retrieval and code-execution steps, while each step also receives localized feedback from its own verifiable reward.

\begin{table*}[t]
\centering
\setlength{\tabcolsep}{2pt}
\resizebox{\textwidth}{!}{
\begin{tabular}{lcccccccccccc}
\toprule
 & \multicolumn{4}{c}{Raman} & \multicolumn{4}{c}{UV-Vis} & \multicolumn{4}{c}{IR} \\
\cmidrule(lr){2-5} \cmidrule(lr){6-9} \cmidrule(lr){10-13}
Model & Tanimoto & Cosine & MACCS & Fraggle & Tanimoto & Cosine & MACCS & Fraggle & Tanimoto & Cosine & MACCS & Fraggle \\
\midrule
\multicolumn{13}{c}{\textit{General LLMs}} \\
GPT-5.5 & 0.1004 & 0.1908 & 0.1797 & 0.0419 & 0.0154 & 0.0424 & 0.0422 & 0.0021 & 0.0489 & 0.0941 & 0.1203 & 0.0449 \\
Gemini-3.1-Pro & 0.0976 & 0.1848 & 0.1950 & 0.0797 & 0.0250 & 0.0607 & 0.0639 & 0.0018 & 0.0331 & 0.0684 & 0.0832 & 0.0197 \\
DeepSeek-V4 Pro & 0.0840 & 0.1697 & 0.1573 & 0.0650 & 0.0244 & 0.0516 & 0.0460 & 0.0044 & 0.0577 & 0.1141 & 0.1294 & 0.0437 \\
Claude-Opus-4.7 & 0.0901 & 0.1765 & 0.1961 & 0.1179 & 0.0426 & 0.0860 & 0.1235 & 0.0783 & 0.0332 & 0.0672 & 0.0932 & 0.0389 \\
\midrule
\multicolumn{13}{c}{\textit{Spectrum-specific Models}} \\
IR-to-Structure & 0.0766 & 0.1395 & 0.1639 & 0.1959 & 0.0728 & 0.1326 & 0.1512 & 0.1837 & 0.0718 & 0.1311 & 0.1585 & 0.1747 \\
Spectra2Structure & 0.1089 & 0.1901 & 0.2388 & \underline{0.2504} & 0.0716 & 0.1313 & 0.1418 & 0.2092 & 0.0965 & 0.1695 & 0.2162 & 0.2308 \\
SpectraLLM & \underline{0.2500} & \underline{0.3786} & \underline{0.5071} & 0.2500 & \underline{0.0790} & \underline{0.1426} & \underline{0.2026} & \underline{0.2100} & \underline{0.1921} & \underline{0.3120} & \underline{0.4330} & \underline{0.3194} \\
S3C-LLM & \textbf{0.3354} & \textbf{0.4541} & \textbf{0.5832} & \textbf{0.5879} & \textbf{0.1381} & \textbf{0.2103} & \textbf{0.2914} & \textbf{0.4385} & \textbf{0.2534} & \textbf{0.3834} & \textbf{0.5073} & \textbf{0.5904} \\
\bottomrule
\end{tabular}
}
\vspace{-5pt}
\caption{Evaluation of general and spectrum-specific models on QM9S single-spectrum tasks (Raman, UV-Vis, and IR). \textbf{Bold} and \underline{underline} indicate the best and second-best performances, respectively.}
\vspace{-5pt}
\label{tab:qm9s-results}
\end{table*}

\section{Experiments}

\subsection{Experimental Setup}

\textbf{Benchmarks.} Following SpectraLLM, we evaluate S3C-LLM on spectrum-to-structure benchmarks across multiple spectroscopic modalities and use the same evaluation splits. The evaluation covers QM9S single-spectrum tasks \citep{zou2023deep}, including Raman, UV-Vis, and IR spectra, the Multimodal Spectroscopic Dataset \citep{alberts2024unraveling}, including \textsuperscript{13}C NMR, \textsuperscript{1}H NMR, HSQC, and MS/MS spectra, and MassSpecGym MS/MS spectra \citep{bushuiev2024massspecgym}. All inputs are converted into peak-level textual representations, and the model is required to generate the molecular SMILES from the given spectra. All teacher-assisted data construction steps, including skill evolution, analysis-code generation, thinking-trace generation and filtering, SFT trajectory construction, and RL trajectory sampling, are restricted to the training split; evaluation molecules and spectra are never used in these steps.

\textbf{Inference.} Given a spectrum input and the tool configuration, S3C-LLM is expected to follow the learned tool-use process: it first calls \texttt{read\_skill} to retrieve modality-specific spectroscopy knowledge, then calls \texttt{run\_code} to execute generated analysis code, and finally analyzes the tool responses to output the SMILES.

\textbf{Baselines.} We include two types of baselines. General LLMs include DeepSeek-V4 Pro \citep{deepseek2026v4}, GPT-5.5 \citep{openai2026gpt55}, Claude-Opus-4.7 \citep{anthropic2026claudeopus47}, and Gemini-3.1-Pro \citep{googledeepmind2026gemini31pro}, which directly generate SMILES from optical-spectrum inputs without spectroscopy-specific training or tools. Spectrum-specific models include IR-to-Structure and Spectra2Structure for QM9S, Spec2Mol \citep{litsa2023spec2mol} and DiffMS \citep{bohde2025diffms} for MS/MS, and SpectraLLM on the corresponding QM9S, NMR, and MS/MS settings, with reported results following SpectraLLM \citep{su2026spectrallm}.

\textbf{Metrics.} We compute molecular similarity metrics with RDKit \citep{rdkit}, including Tanimoto, cosine, MACCS Tanimoto, and Fraggle similarities. Since a single spectrum may not uniquely determine one exact molecule, we focus on structure-level similarity rather than exact-match accuracy.

\subsection{Implementation details.} We initialize S3C-LLM from the Qwen3-4B backbone \citep{yang2025qwen3} and SFT it on our constructed 500K agentic trajectories with \texttt{read\_skill} and \texttt{run\_code} tools. Notably, our SFT data is less than 1/10th of the paired-spectra corpus reported by SpectraLLM. For step-level RL, we sample 8 rollouts per prompt with temperature 1.0, use $\gamma=0.95$, and a learning rate of $1\times10^{-6}$, without an explicit KL penalty. For the SFT and RL stages, we perform full-parameter fine-tuning with global batch sizes of 64 and 128, respectively. Both SFT and RL stages are trained for a single epoch on 8 NVIDIA H800 GPUs with 16 and 20 hours, respectively.

\subsection{Main Results}

\textbf{Optical spectra.} We first evaluate S3C-LLM on Raman, UV-Vis, and IR spectra from QM9S. Since each single-spectrum modality in QM9S exposes only partial structural evidence, this setting tests whether the model can infer molecular structures from limited spectral cues. The results in Table~\ref{tab:qm9s-results} show that S3C-LLM performs best across all three single-spectrum settings compared with both modality-specific baselines and SpectraLLM. These results demonstrate the data efficiency of S3C-LLM, as we use less than 1/10th of the paired-spectra corpus reported by SpectraLLM, and suggest that our agentic model can effectively exploit peak information through skills and code.

\textbf{Mass spectra.} We then evaluate S3C-LLM on MS/MS spectra from the Multimodal Spectroscopic Dataset \citep{alberts2024unraveling} and MassSpecGym \citep{bushuiev2024massspecgym}. Since mass spectra provide molecular-mass and fragmentation evidence, this setting evaluates whether the model can convert mass constraints into plausible molecular formulas and structures. In S3C-LLM, the MS/MS skills describe how fragment-ion peaks, neutral-loss patterns, and mass differences constrain candidate formulas, while code execution enumerates and ranks formula combinations that match the observed masses. For larger molecules with many possible formulas, the code returns top-$k$ formula candidates, and the model further reasons over the spectrum to predict the final structure. The results in Table~\ref{tab:msms-results} show that S3C-LLM achieves superior performance on both MS/MS benchmarks, with especially large gains on the Multimodal Spectroscopic Dataset. This is expected because its simulated spectra provide cleaner mass constraints for code-based formula enumeration, whereas experimental MassSpecGym spectra contain measurement noise and peak shifts that make mass matching less exact.

\begin{table}[t]
\centering
\setlength{\tabcolsep}{3pt}
\resizebox{\linewidth}{!}{
\begin{tabular}{lccccc}
\toprule
Method & Tanimoto & Cosine & FG & MACCS & Fraggle \\
\midrule
\multicolumn{6}{c}{\textit{Multimodal Spectroscopic Dataset}} \\
Spec2Mol & 0.0988 & 0.1739 & 0.3042 & 0.2440 & 0.2587 \\
DiffMS & 0.1535 & 0.2351 & 0.4248 & 0.3730 & 0.3635 \\
SpectraLLM & \underline{0.1844} & \underline{0.2993} & \underline{0.4929} & \underline{0.4254} & \underline{0.4282} \\
S3C-LLM & \textbf{0.6382} & \textbf{0.7027} & \textbf{0.7519} & \textbf{0.7749} & \textbf{0.6563} \\
\midrule
\multicolumn{6}{c}{\textit{MassSpecGym}} \\
Spec2Mol & 0.0849 & 0.1511 & 0.3111 & 0.2709 & 0.2065 \\
DiffMS & \underline{0.1597} & 0.2422 & 0.4890 & 0.4305 & 0.3539 \\
SpectraLLM & 0.1533 & \underline{0.2558} & \underline{0.5003} & \underline{0.4723} & \underline{0.3610} \\
S3C-LLM & \textbf{0.1832} & \textbf{0.2974} & \textbf{0.5219} & \textbf{0.5485} & \textbf{0.3961} \\
\bottomrule
\end{tabular}
}
\vspace{-5pt}
\caption{MS/MS spectrum-to-structure results on the Multimodal Spectroscopic Dataset and MassSpecGym.}
\vspace{-5pt}
\label{tab:msms-results}
\end{table}

\textbf{NMR spectra.} Table~\ref{tab:nmr-results} evaluates S3C-LLM on \textsuperscript{13}C NMR, \textsuperscript{1}H NMR, HSQC, and their combined setting from the Multimodal Spectroscopic Dataset. These NMR modalities provide different structural cues: carbon skeleton information, local proton environments, and proton-carbon cross-peak correlations. The single-spectrum results show that S3C-LLM consistently outperforms SpectraLLM across all single-NMR settings. This further demonstrates that S3C-LLM is not limited to vibrational or mass spectra, but can also generalize to chemical-shift and cross-peak representations. Following the multi-spectral evaluation in SpectraLLM, we also evaluate the combined NMR setting with \textsuperscript{13}C NMR, \textsuperscript{1}H NMR, and HSQC spectra. The results show that S3C-LLM achieves the best performance by exploiting complementary NMR evidence.

\begin{table}[t]
\centering
\setlength{\tabcolsep}{3pt}
\resizebox{\linewidth}{!}{
\begin{tabular}{lcccc}
\toprule
Model & Tanimoto $\uparrow$ & Cosine $\uparrow$ & MACCS $\uparrow$ & Fraggle $\uparrow$ \\
\midrule
\multicolumn{5}{c}{\textit{\textsuperscript{13}C NMR}} \\
SpectraLLM & \underline{0.1016} & \underline{0.1801} & \underline{0.2952} & \underline{0.3607} \\
S3C-LLM & \textbf{0.2508} & \textbf{0.3791} & \textbf{0.5239} & \textbf{0.4846} \\
\midrule
\multicolumn{5}{c}{\textit{\textsuperscript{1}H NMR}} \\
SpectraLLM & \underline{0.0720} & \underline{0.1341} & \underline{0.2203} & \underline{0.2572} \\
S3C-LLM & \textbf{0.2281} & \textbf{0.3577} & \textbf{0.5002} & \textbf{0.4456} \\
\midrule
\multicolumn{5}{c}{\textit{HSQC}} \\
SpectraLLM & \underline{0.2058} & \underline{0.3221} & \underline{0.4392} & \underline{0.4274} \\
S3C-LLM & \textbf{0.4016} & \textbf{0.5312} & \textbf{0.6491} & \textbf{0.5965} \\
\midrule
\multicolumn{5}{c}{\textit{\textsuperscript{13}C NMR + \textsuperscript{1}H NMR + HSQC}} \\
NMR2Struct & 0.0433 & 0.1029 & 0.1294 & 0.0962 \\
SpectraLLM & \underline{0.4151} & \underline{0.5322} & \underline{0.6367} & \underline{0.5862} \\
S3C-LLM & \textbf{0.5021} & \textbf{0.5895} & \textbf{0.6743} & \textbf{0.6013} \\
\bottomrule
\end{tabular}
}
\vspace{-5pt}
\caption{NMR spectrum-to-structure results on the Multimodal Spectroscopic Dataset, including single-NMR and combined-NMR settings.}
\label{tab:nmr-results}
\end{table}

\subsection{Ablation Study}

We analyze S3C-LLM from three complementary aspects: the reliability of the evolved skills, the contribution of each training stage, and the effect of different tool configurations.

\begin{figure}[t]
\centering
\includegraphics[width=0.98\linewidth]{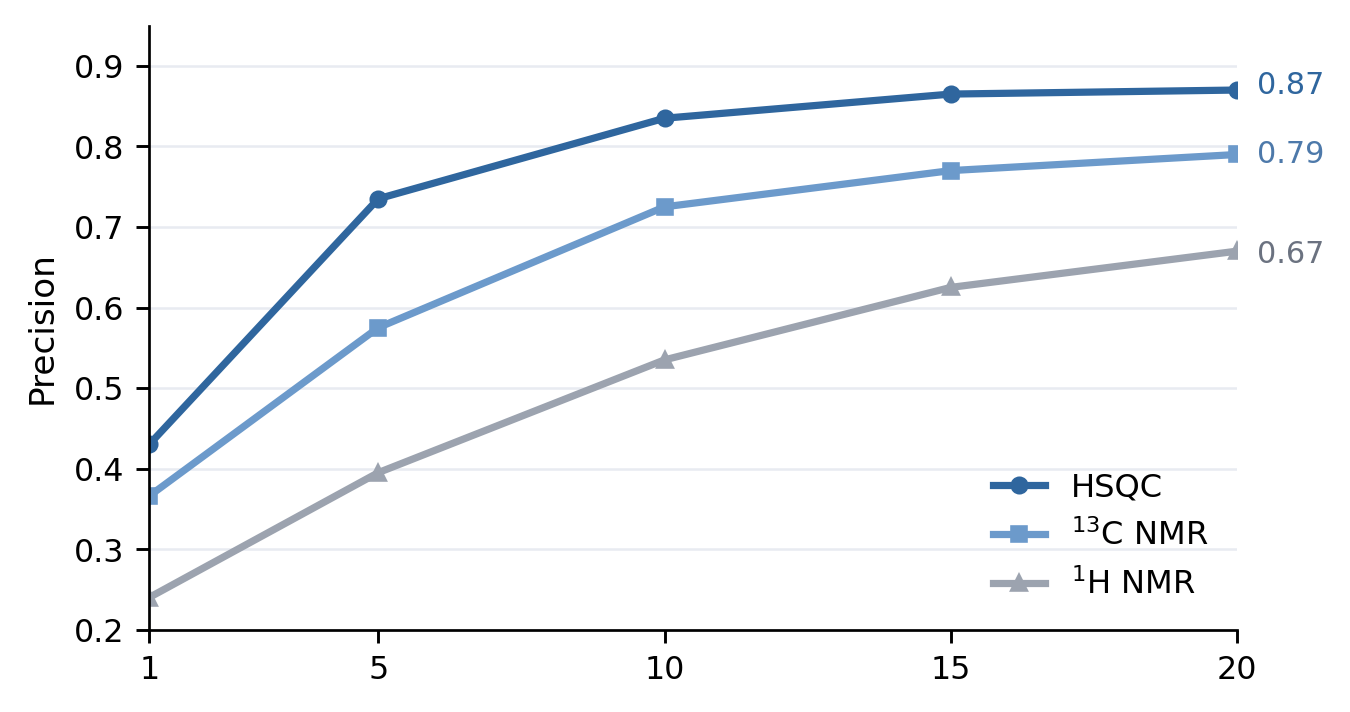}
\vspace{-5pt}
\caption{Skill quality across evolution rounds, measured by precision on 50 training-split spectra checked in each round.}
\vspace{-5pt}
\label{fig:skill-quality}
\end{figure}

\begin{table}[t]
\centering
\setlength{\tabcolsep}{3pt}
\resizebox{\linewidth}{!}{
\begin{tabular}{lcccc}
\toprule
Variant & \textsuperscript{13}C NMR & \textsuperscript{1}H NMR & HSQC & MS/MS \\
\midrule
Only SFT & 0.2069 & \underline{0.1914} & 0.3580 & 0.5542 \\
Only Step RL & 0.0740 & 0.0310 & 0.0850 & 0.1310 \\
SFT + Traj. RL & \underline{0.2231} & 0.1847 & \underline{0.3632} & \underline{0.5932} \\
SFT + Step RL & \textbf{0.2508} & \textbf{0.2281} & \textbf{0.4016} & \textbf{0.6382} \\
\bottomrule
\end{tabular}
}
\vspace{-5pt}
\caption{Training-stage ablation across NMR and MS/MS settings. Values are Tanimoto similarity.}
\vspace{-5pt}
\label{tab:training-stages}
\end{table}

\subsubsection{Skill Quality Analysis}
We first evaluate the quality of the constructed spectroscopy skills before testing the trained model. This analysis is important because S3C-LLM relies on skills as reusable spectral priors, and noisy peak-to-structure rules would propagate errors into trajectory construction and model training. During skill evolution, we use a conservative writing criterion: only deterministic or high-confidence spectral rules are added to the skill library, while ambiguous peak-to-structure associations are left out. Therefore, the skills are not expected to infer every functional group in the target SMILES from the spectrum alone; instead, they provide reliable reference evidence and chemical constraints for subsequent reasoning.

For each evolution round, we sample 50 spectra from the training split to check the current skills and compute precision over explicit skill-supported claims. A claim is counted as correct when its peak interpretation, functional-group hypothesis, or chemical constraint is consistent with the reference structure. Ambiguous cues that are not written into the skill are not counted as errors. This precision is therefore high because it evaluates the correctness of stated skill evidence rather than recall over all functional groups in the target SMILES, which remains substantially more underdetermined. As shown in Figure~\ref{fig:skill-quality}, skill evolution steadily improves precision across all three NMR modalities. HSQC converges earlier because cross-peaks provide more explicit structural evidence, while \textsuperscript{13}C NMR and \textsuperscript{1}H NMR improve more gradually.

\subsubsection{Training Stage Ablation}

We next ablate the training stages of S3C-LLM on three NMR settings and the Multimodal MS/MS setting. As shown in Table~\ref{tab:training-stages}, SFT provides the basic agentic tool-use ability, while step-level RL further optimizes the SFT model to form the default S3C-LLM and consistently improves performance across all spectrum types. In contrast, applying step-level RL directly to Qwen3-4B without SFT brings almost no effective learning, because the model has not yet learned the tool-use format and the spectrum-analysis procedure. Replacing step-level credit assignment with trajectory-level RL gives smaller and less stable gains, suggesting that localized rewards for skill reading, code execution, and final prediction are important for training agentic spectroscopy models.

\subsubsection{Toolset Ablation}

\begin{figure}[t]
\centering
\includegraphics[width=0.98\linewidth]{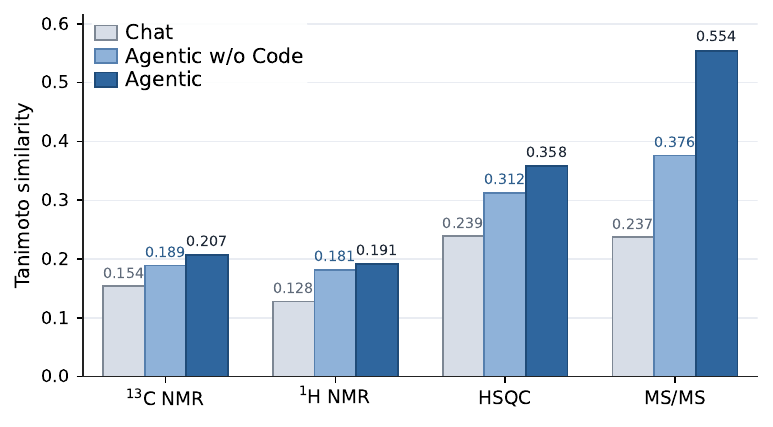}
\vspace{-5pt}
\caption{Toolset ablation with Chat, Agentic w/o Code, and Agentic settings. Bars report Tanimoto similarity.}
\vspace{-5pt}
\label{fig:skill-code-ablation}
\end{figure}

All variants in this ablation use the Qwen3-4B backbone and the same spectrum-structure training instances, while the interaction format and available tools differ. The default S3C-LLM setting uses two tools, \texttt{read\_skill} and \texttt{run\_code}, so that the model can retrieve spectroscopy skills and execute code during spectrum analysis. We compare it with two controlled variants: a chat-format model trained without any tool configuration, and an agentic model trained with only the \texttt{read\_skill} tool. As shown in Figure~\ref{fig:skill-code-ablation}, adding \texttt{read\_skill} improves over the chat baseline across all evaluated spectra, showing that explicit spectral priors help the model interpret peak evidence. Adding \texttt{run\_code} further improves performance, especially on HSQC and MS/MS, where cross-peak relations and mass constraints benefit more from executable computation.

\section{Conclusion}

We present S3C-LLM, a skill-code guided agentic LLM for spectrum-to-structure elucidation. By combining spectroscopy skills with executable code, S3C-LLM reasons over peak evidence, formula constraints, and intermediate computations before predicting structures. Experiments across optical, MS/MS, and NMR spectra show that S3C-LLM outperforms modality-specific baselines and SpectraLLM while using a smaller Qwen3-4B backbone and substantially fewer training samples. Ablations confirm the value of agentic training, step-level RL, retrieved skills, and code execution. Future work will extend S3C-LLM to noisier experimental spectra.

\section*{Limitations}

S3C-LLM is evaluated on established spectrum-to-structure benchmarks using compact peak-level representations, which follows prior work and enables fair comparison across methods. Future work can further extend the skill-guided and code-grounded framework to richer raw-spectrum inputs and additional experimental conditions. The current spectroscopy skill library covers common optical, NMR, and MS/MS modalities, and can be expanded to more specialized spectra as new datasets become available. Since S3C-LLM performs inference through tool-use trajectories, improving tool-call scheduling and execution efficiency is another useful direction for large-scale deployment.

\section*{Ethics Statement}

This work uses public or benchmark spectroscopy datasets and does not involve human subjects or private personal data. The intended use of S3C-LLM is scientific support for molecular structure elucidation. Since structure prediction systems may be connected to molecule design workflows, outputs should not be used directly for hazardous synthesis, controlled-substance discovery, or safety-critical chemical decisions without appropriate screening and human oversight. We do not provide wet-lab protocols or optimize molecules for biological activity or toxicity.

\bibliographystyle{acl_natbib}
\bibliography{custom}

\clearpage
\appendix
\section{Appendix}
\subsection{Dataset and Split Details}

We use the same benchmark sources described in Section~4.1 and do not introduce additional evaluation splits. Following SpectraLLM, the evaluation covers QM9S single-spectrum tasks with Raman, UV-Vis, and IR spectra; the Multimodal Spectroscopic Dataset with \textsuperscript{13}C NMR, \textsuperscript{1}H NMR, HSQC, MS/MS, and combined-NMR settings; and MassSpecGym for experimental MS/MS spectra. All spectra are converted into compact peak-level textual representations before being used as model inputs.

All data-construction stages that use reference structures are restricted to the training split. This includes spectroscopy skill evolution, analysis-code generation, thinking-trace generation and filtering, SFT trajectory construction, and RL trajectory sampling. Evaluation molecules and spectra are not used in these teacher-assisted construction stages. The held-out benchmark splits are used only for final evaluation, where S3C-LLM receives the spectrum input and tool configuration, performs the learned \texttt{read\_skill} and \texttt{run\_code} tool-use process, and outputs the final SMILES.


\subsection{Comparison with SpectraLLM}

S3C-LLM and SpectraLLM differ in how they use model scale and supervised data. SpectraLLM adopts a larger Qwen3-32B backbone and fine-tunes it with LoRA on direct spectrum-to-SMILES examples. Its constructed corpus combines multiple spectral datasets and is reported to contain more than 5.5 million paired spectra. In contrast, S3C-LLM uses a smaller Qwen3-4B backbone and trains on 500K skill-code trajectories, where the model learns to retrieve spectroscopy skills, execute analysis code, and then predict the final SMILES.

\subsection{Skill Examples}

Figures~\ref{fig:appendix-hsqc-skill} and~\ref{fig:appendix-msms-skill} show the complete Markdown skill files used by \texttt{read\_skill}. These examples are selected from the HSQC and MS/MS skills and illustrate how the skill library stores reusable spectral priors for downstream code-grounded reasoning.

\subsection{Correct Rollout Trajectories}

Figures~\ref{fig:appendix-rollout-hsqc}--\ref{fig:appendix-rollout-c-nmr} show three successful tool-use trajectories where the final predicted SMILES exactly matches the reference structure. Each box preserves the conversation trajectory itself, including user input, assistant reasoning, tool calls, tool responses, code observations, and final prediction.

\clearpage
\onecolumn

\begin{tcblisting}{rolloutlisting,title={\texttt{skill\_hsqc.md}}}
# HSQC (Heteronuclear Single Quantum Coherence) NMR Skill

## What it measures
HSQC is a 2D NMR experiment that correlates each 1H signal with the directly bonded 13C. Each cross-peak represents a
C-H bond. Quaternary carbons (no H) are invisible.

## Key observables
- **1H chemical shift (ppm)**: Hydrogen environment
- **13C chemical shift (ppm)**: Carbon environment
- **nH**: Number of H on that carbon (1=CH, 2=CH2, 3=CH3)

## Combined (1H, 13C) -> precise assignment

The power of HSQC is that combining both dimensions resolves ambiguities:

| 1H (ppm) | 13C (ppm) | nH | Assignment |
|-----------|-----------|-----|-----------|
| 0.0-1.0 | -5-25 | 3 | R-CH3/cyPr (alkyl CH3 or cyclopropyl) |
| 1.0-2.0 | 0-25 | 3 | R-CH3 (methyl) |
| 1.0-2.0 | 25-50 | 2 | R-CH2-R (alkyl CH2) |
| 0.5-1.8 | 25-50 | 2 | R-CH2-R (alkyl CH2) |
| 1.8-3.0 | 0-25 | 3 | =C-CH3/S-CH3 (allyl or thioether methyl) |
| 2.0-3.0 | 20-50 | 2-3 | C=O-CH/allyl (alpha to carbonyl) |
| 2.0-3.5 | 25-65 | 2-3 | N-CH/S-CH (bonded to N or S) |
| 3.0-4.5 | 20-50 | 2 | N-CH2/N-CH (nitrogen methylene) |
| 3.0-4.5 | 50-90 | 3 | O-CH(sp3) (methoxy, alcohol, ether) |
| 4.0-5.0 | 55-80 | 2 | O-CH2(ester) |
| 4.5-5.5 | 30-60 | 1 | N-CH/act-CH (activated CH) |
| 4.5-6.0 | 85-110 | 1 | O-CH-O (anomeric, acetal) |
| 5.0-6.5 | 60-85 | 1 | O-CH(allyl) |
| 5.0-7.0 | 95-140 | 1 | =CH(vinyl) (olefinic) |
| 6.0-7.0 | 80-120 | 1 | Ar-H/HetAr (aromatic/heteroaromatic) |
| 6.5-8.0 | 100-145 | 1 | Ar-H (aromatic) |
| 7.5-9.0 | 100-140 | 1 | Ar-H(EWG) (electron-withdrawing group adjacent) |
| 7.0-9.0 | 140-165 | 1 | Ar-H(subst) (deshielded, C-N/C-O adjacent) |
| 8.0-9.5 | 115-165 | 1 | Ar-H/HetAr(ds) (highly deshielded) |
| 9.0-10.5 | 175-220 | 1 | CHO (aldehyde) |
| 8.0-8.5 | 160-175 | 1 | HCO(formate) (formate/formamide) |
| 2.0-3.5 | 60-90 | 1 | C#C-CH/epoxy (propargylic or epoxide) |
| 3.0-3.5 | 0-25 | 3 | cyPr/N-CH3 (cyclopropyl or N-methyl) |
| 4.5-5.0 | 75-90 | 1 | O-CH(sugar) (sugar-type) |
| 5.0-6.0 | 40-60 | 1 | =CH(strained) (strained olefinic) |
| 6.0-7.5 | 55-100 | 1 | HetAr/=CH(ds) (heteroaromatic, deshielded vinyl) |
| 7.0-8.0 | 90-100 | 1 | HetAr(5-ring) (five-membered heteroaromatic) |

> Note: Ranges deliberately overlap. A correlation in an overlap zone matches multiple groups - all possibilities are reported.

## CH multiplicity from nH

| nH | Carbon type | What it tells you |
|----|------------|-------------------|
| 3 | CH3 | Methyl group. If at ~3.8/55 ppm -> OCH3. If at ~0.9/14 ppm -> terminal CH3. |
| 2 | CH2 | Methylene. Count these to estimate chain length. |
| 1 | CH | Methine or aromatic. If in aromatic region -> aromatic H count. |
| 0 | Quaternary C | Not visible in HSQC (no cross-peak). |

## Structural reasoning

### Aromatic ring detection
- Cross-peaks at (6.5-8.5, 110-145) with nH=1 -> aromatic CH
- Count of aromatic CH correlations -> substitution pattern
  - 5 aromatic CH -> monosubstituted benzene
  - 4 aromatic CH -> disubstituted benzene
  - 2-3 aromatic CH -> tri/tetrasubstituted

### Functional group counting
- Total H from HSQC = sum(nH for all cross-peaks)
- Missing H vs molecular formula -> NH, OH (exchangeable, not in HSQC)
- No cross-peak at certain 13C -> quaternary carbon (C=O, C-quaternary)

### OCH3 fingerprint
- (3.8-4.0, 53-58, nH=3) is almost always OCH3 (methoxy)
- Very diagnostic: if you see this, molecule has -OCH3
\end{tcblisting}
\listingcaption{Full Markdown content of the HSQC skill used by S3C-LLM.}{fig:appendix-hsqc-skill}

\clearpage
\begin{tcblisting}{rolloutlisting,title={\texttt{skill\_msms.md}}}
# MS/MS (Tandem Mass Spectrometry) Skill

## What it measures
MS/MS measures molecular fragmentation. A precursor ion is selected and fragmented by collision, producing fragment
ions. The pattern of fragments reveals the molecular structure.

## Key observables
- **Precursor m/z**: Molecular ion mass (gives molecular weight)
- **Adduct type**: [M+H]+ (positive, M = precursor - 1.0073), [M+Na]+ (M = precursor - 22.9892), [M-H]- (negative, M = precursor + 1.0073)
- **Fragment m/z values**: Masses of breakdown products
- **Fragment intensity**: Relative abundance (higher = more favorable fragmentation)
- **Fragment annotations** (if available): Molecular formula or SMILES of each fragment

## Step 1: Molecular formula from precursor mass

Given the neutral mass M, enumerate possible molecular formulas CxHyNnOoSs... within mass tolerance (typically 5 ppm for high-resolution MS):
- **Small molecules (<250 Da)**: Few candidates, usually unique at 5 ppm
- **Medium molecules (250-500 Da)**: ~10 candidates, need element constraints
- **Large molecules (>500 Da)**: Many candidates, use 10 ppm, fragment annotations help constrain elements

### Element constraints from fragments
If fragment annotations are molecular formulas (e.g., C4O3H4), the union of elements across all fragments tells you
which elements exist in the molecule. This dramatically reduces the search space.

### RDBE (Ring and Double Bond Equivalence)
RDBE = 1 + C - H/2 + N/2 (for CHNO molecules)
- RDBE < 0 -> chemically impossible
- RDBE = 4 -> one benzene ring or equivalent
- RDBE >= 4 with aromatic fragments -> aromatic compound

### Chemical plausibility scoring for formula ranking
When multiple formula candidates have similar ppm errors, rank by plausibility:
- **N/C ratio > 0.4** -> penalty (most organics have N/C < 0.4)
- **O/C ratio > 1.0** -> penalty (most organics have O/C < 1.0)
- **RDBE/C ratio > 0.65** -> penalty (overly unsaturated; benzene = 0.67 is the practical limit for typical organics)
- **RDBE > mass/20** -> penalty (too many degrees of unsaturation for the molecular size)
- **C < mass/20** -> penalty (too few carbons for the molecular weight)
- Formulas with no carbon but containing N/O -> strongly penalized

## Step 2: Neutral loss analysis

Neutral loss = precursor m/z - fragment m/z. Each loss corresponds to a specific structural moiety:

| Loss (Da) | Formula | Functional group |
|-----------|---------|-----------------|
| 15.024 | CH3 | Methyl group |
| 17.003 | OH | Hydroxyl radical |
| 18.011 | H2O | Alcohol, carboxylic acid (dehydration) |
| 27.995 | CO | Carbonyl group (ketone, quinone) |
| 28.031 | C2H4 | Ethyl group (ethyl ester/ether retro-cleavage) |
| 31.018 | OCH3 | Methoxy group |
| 32.026 | CH3OH | Methanol (methyl ester) |
| 42.011 | CH2CO | Ketene (acetyl group) |
| 43.990 | CO2 | Carboxyl group (-COOH) |
| 46.005 | CH2O2 | Formic acid (formate ester) |
| 60.021 | C2H4O2 | Acetic acid (acetate ester) |
| 63.962 | SO2 | Sulfone, sulfonate |
| 79.957 | SO3 | Sulfate ester |
| 79.966 | HPO3 | Phosphate |
| 162.053 | C6H10O5 | Hexose sugar (glycosides) |
| 176.032 | C6H8O6 | Glucuronic acid |

### Reasoning with neutral losses
- Loss of 18 (H2O) -> molecule has OH group (alcohol, carboxylic acid, or hemiacetal)
- Loss of 28 (CO) -> carbonyl present, often from ketones/quinones after ring opening
- Loss of 44 (CO2) -> carboxylic acid or carbonate
- Loss of 31 (OCH3) -> methoxy group present
- Sequential losses: H2O + CO = loss of 46 -> phenolic acid fragmentation

## Step 3: Diagnostic fragment ions

Certain fragment m/z values are structural fingerprints. **Caution**: matching by m/z alone can cause false positives -
a peak near m/z 91 does not guarantee tropylium. Always cross-check with neutral losses and molecular formula.

| m/z | Ion | Structural meaning | False positive risk |
|-----|-----|-------------------|-------------------|
| 77.04 | C6H5+ | Phenyl cation -> benzene ring | Low - very specific |
| 91.05 | C7H7+ | Tropylium/benzyl -> toluene/benzyl moiety | **Medium** - other C7H7 isomers exist |
| 105.03 | C7H5O+ | Benzoyl -> benzoic acid derivative | Low |
| 120.08 | C8H10N+ | Phenylalanine immonium -> contains Phe | **High** if not a peptide |
| 130.07 | C9H8N+ | Tryptophan immonium -> contains Trp | **High** if not a peptide |
| 136.06 | C5H6N5+ | Adenine+H -> nucleoside/nucleotide | **High** if not nucleotide |
| 147.04 | C9H7O2+ | Coumarin fragment -> coumarin scaffold | Medium |
| 152.06 | C5H6N5O+ | Guanine+H -> guanine nucleoside | **High** if not nucleotide |

**Matching tolerance**: Use +/-0.003 Da (3 mDa) for high-resolution data. Wider tolerances cause frequent false
positives - e.g., a non-peptide fragment at m/z 130.065 can be misidentified as tryptophan immonium (130.066). When in
doubt, cross-validate with neutral losses.

## Data source differences

| Source | Fragment annotation | Precursor type | Notes |
|--------|-------------------|---------------|-------|
| ICEBERG | Molecular formula (C4O3H4) | [M+H]+ (ion m/z) | ~300 fragments per spectrum |
| SCARF | Molecular formula | Neutral mass M | ~300 fragments |
| CFM-ID (pos) | SMILES fragment | [M+H]+ (ion m/z) | ~10-40 fragments |
| CFM-ID (neg) | SMILES fragment | [M-H]- (ion m/z) | ~3-25 fragments |
| MassSpecGym | None (m/z + intensity only) | [M+H]+ or [M+Na]+ (explicit in metadata) | Real experimental data |

## MassSpecGym limitations
MassSpecGym has **no fragment annotations**, so:
- Element detection relies on default assumption (C,H,N,O,S) - may miss halogens (F,Cl,Br)
- Formula candidates are less constrained -> top-1 accuracy is lower for medium/large molecules
- Formula ranking should be treated as a candidate list, not a definitive answer
- Neutral loss and diagnostic fragment analysis remain valid regardless
\end{tcblisting}
\listingcaption{Full Markdown content of the MS/MS skill used by S3C-LLM.}{fig:appendix-msms-skill}

\clearpage
\begin{tcblisting}{rolloutlisting,title={Correct HSQC Rollout}}
[1] USER
--------
Given Heteronuclear Single Quantum Coherence {"H-shifts": "3.88,3.37,0.93,1.22,7.28,3.7,7.9,1.67,1.42", "C-shifts":
"52.27,41.04,10.27,20.04,109.89,49.34,114.29,29.81,29.81", "Intensities": "3.0,3.0,3.0,3.0,1.0,1.0,1.0,1.0,1.0"}, the
spectra data includes the Chemical Shift positions in ppm as C-shifts,H-shifts and corresponding intensities as
Intensities. Based on the information provided, predict which compound the spectra correspond to and give the SMILES of
that compound. Please answer strictly in the format ##SMILES: .

[2] ASSISTANT
-------------
<think>
This HSQC spectrum shows aromatic correlations between 109-114 ppm and aliphatic signals in the 10-52 ppm range. I will
load the HSQC skill to interpret these one-bond carbon-proton couplings.
</think><tool_call>
{"name": "read_skill", "arguments": {"name": "hsqc"}}
</tool_call>

[3] TOOL
-----------------------------------------------------------------
# HSQC (Heteronuclear Single Quantum Coherence) NMR Skill

## What it measures
HSQC is a 2D NMR experiment that correlates each 1H signal with the directly bonded 13C. Each cross-peak represents a
C-H bond. Quaternary carbons (no H) are invisible.

## Key observables
- **1H chemical shift (ppm)**: Hydrogen environment
- **13C chemical shift (ppm)**: Carbon environment
- **nH**: Number of H on that carbon (1=CH, 2=CH2, 3=CH3)

## Combined (1H, 13C) -> precise assignment

The power of HSQC is that combining both dimensions resolves ambiguities:

| 1H (ppm) | 13C (ppm) | nH | Assignment |
|-----------|-----------|-----|-----------|
| 0.0-1.0 | -5-25 | 3 | R-CH3/cyPr (alkyl CH3 or cyclopropyl) |
| 1.0-2.0 | 0-25 | 3 | R-CH3 (methyl) |
| 1.0-2.0 | 25-50 | 2 | R-CH2-R (alkyl CH2) |
| 0.5-1.8 | 25-50 | 2 | R-CH2-R (alkyl CH2) |
| 1.8-3.0 | 0-25 | 3 | =C-CH3/S-CH3 (allyl or thioether methyl) |
| 2.0-3.0 | 20-50 | 2-3 | C=O-CH/allyl (alpha to carbonyl) |
| 2.0-3.5 | 25-65 | 2-3 | N-CH/S-CH (bonded to N or S) |
| 3.0-4.5 | 20-50 | 2 | N-CH2/N-CH (nitrogen methylene) |
| 3.0-4.5 | 50-90 | 3 | O-CH(sp3) (methoxy, alcohol, ether) |
| 4.0-5.0 | 55-80 | 2 | O-CH2(ester) |
| 4.5-5.5 | 30-60 | 1 | N-CH/act-CH (activated CH) |
| 4.5-6.0 | 85-110 | 1 | O-CH-O (anomeric, acetal) |
| 5.0-6.5 | 60-85 | 1 | O-CH(allyl) |
| 5.0-7.0 | 95-140 | 1 | =CH(vinyl) (olefinic) |
| 6.0-7.0 | 80-120 | 1 | Ar-H/HetAr (aromatic/heteroaromatic) |
| 6.5-8.0 | 100-145 | 1 | Ar-H (aromatic) |
| 7.5-9.0 | 100-140 | 1 | Ar-H(EWG) (electron-withdrawing group adjacent) |
| 7.0-9.0 | 140-165 | 1 | Ar-H(subst) (deshielded, C-N/C-O adjacent) |
| 8.0-9.5 | 115-165 | 1 | Ar-H/HetAr(ds) (highly deshielded) |
| 9.0-10.5 | 175-220 | 1 | CHO (aldehyde) |
| 8.0-8.5 | 160-175 | 1 | HCO(formate) (formate/formamide) |
| 2.0-3.5 | 60-90 | 1 | C#C-CH/epoxy (propargylic or epoxide) |
| 3.0-3.5 | 0-25 | 3 | cyPr/N-CH3 (cyclopropyl or N-methyl) |
| 4.5-5.0 | 75-90 | 1 | O-CH(sugar) (sugar-type) |
| 5.0-6.0 | 40-60 | 1 | =CH(strained) (strained olefinic) |
| 6.0-7.5 | 55-100 | 1 | HetAr/=CH(ds) (heteroaromatic, deshielded vinyl) |
| 7.0-8.0 | 90-100 | 1 | HetAr(5-ring) (five-membered heteroaromatic) |

> Note: Ranges deliberately overlap. A correlation in an overlap zone matches multiple groups - all possibilities are reported.

## CH multiplicity from nH

| nH | Carbon type | What it tells you |
|----|------------|-------------------|
| 3 | CH3 | Methyl group. If at ~3.8/55 ppm -> OCH3. If at ~0.9/14 ppm -> terminal CH3. |
| 2 | CH2 | Methylene. Count these to estimate chain length. |
| 1 | CH | Methine or aromatic. If in aromatic region -> aromatic H count. |
| 0 | Quaternary C | Not visible in HSQC (no cross-peak). |

## Structural reasoning

### Aromatic ring detection
- Cross-peaks at (6.5-8.5, 110-145) with nH=1 -> aromatic CH
- Count of aromatic CH correlations -> substitution pattern
  - 5 aromatic CH -> monosubstituted benzene
  - 4 aromatic CH -> disubstituted benzene
  - 2-3 aromatic CH -> tri/tetrasubstituted

### Functional group counting
- Total H from HSQC = sum(nH for all cross-peaks)
- Missing H vs molecular formula -> NH, OH (exchangeable, not in HSQC)
- No cross-peak at certain 13C -> quaternary carbon (C=O, C-quaternary)

### OCH3 fingerprint
- (3.8-4.0, 53-58, nH=3) is almost always OCH3 (methoxy)
- Very diagnostic: if you see this, molecule has -OCH3

[4] ASSISTANT
-------------
<think>
Got the hsqc rules. Let me run the code to classify each peak.
</think><tool_call>
name: run_code
arguments:
  code: |
    RULES = [(0.0, 1.0, -5.0, 25.0, 'R-CH3/cyPr'), (1.0, 2.0, 0.0, 25.0, 'R-CH3'), (1.0, 2.0, 25.0, 50.0, 'R-CH2-R'),
    (2.0, 3.5, 25.0, 65.0, 'N-CH/S-CH'), (3.0, 4.5, 50.0, 90.0, 'O-CH(sp3)'), (3.0, 4.5, 20.0, 50.0, 'N-CH2/N-CH'),
    (6.5, 8.0, 100.0, 145.0, 'Ar-H'), (7.5, 9.0, 100.0, 140.0, 'Ar-H(EWG)')]
    correlations = [(3.88, 52.3, 3.0), (3.37, 41.0, 3.0), (0.93, 10.3, 3.0), (1.22, 20.0, 3.0), (7.28, 109.9, 1.0),
    (3.7, 49.3, 1.0), (7.9, 114.3, 1.0), (1.67, 29.8, 1.0), (1.42, 29.8, 1.0)]
    total_h = sum(int(nH) for _, _, nH in correlations if nH)
    print("HSQC 2D NMR Analysis")
    print(f"Correlations: {len(correlations)} | Total H: {total_h}")
    summary = {}
    ch3=ch2=ch=0
    ar_h=0
    fingerprints=[]
    for h, c, nH in correlations:
        nh = int(nH) if nH else 1
        matches = [group for h_lo, h_hi, c_lo, c_hi, group in RULES if h_lo <= h < h_hi and c_lo <= c < c_hi]
        if not matches:
            matches = ["Unknown"]
        for g in matches:
            summary[g] = summary.get(g, 0) + nh
        if nh == 3: ch3 += 1
        elif nh == 2: ch2 += 1
        elif nh == 1: ch += 1
        if 6.0 < h < 9.5 and 90 < c < 165:
            ar_h += nh
        if 3.5 < h < 4.2 and 50 < c < 60 and nh == 3:
            fingerprints.append(f"OCH3 or NCH3 detected (1H={h:.1f}, 13C={c:.1f})")
        if 2.2 < h < 3.0 and 28 < c < 45 and nh == 3:
            fingerprints.append(f"N-CH3 detected (1H={h:.1f}, 13C={c:.1f})")
        if 4.3 < h < 5.8 and 90 < c < 110 and nh == 1:
            fingerprints.append(f"Anomeric CH detected (1H={h:.1f}, 13C={c:.1f})")
    print("H count by group:")
    for g, c in sorted(summary.items(), key=lambda x: -x[1]):
        print(f"  {g}: {c}H")
    print(f"CH classification: CH3 x{ch3}, CH2 x{ch2}, CH x{ch}")
    if ar_h > 0:
        sub = {1:"penta-sub",2:"tetra-sub",3:"tri-sub",4:"di-sub",5:"mono-sub"}
        pat = sub.get(ar_h, f"{ar_h} Ar-H")
        print(f"Aromatic H: {ar_h} ({pat} benzene or polyaromatic)")
    if fingerprints:
        print("Fingerprints:")
        for fp in dict.fromkeys(fingerprints):
            print(f"  {fp}")
</tool_call>

[5] TOOL
-----------------------------------------------------------------
HSQC 2D NMR Analysis
Correlations: 9 | Total H: 17
H count by group:
  N-CH2/N-CH: 4H
  O-CH(sp3): 3H
  N-CH/S-CH: 3H
  R-CH3/cyPr: 3H
  R-CH3: 3H
  Ar-H: 2H
  R-CH2-R: 2H
  Ar-H(EWG): 1H
CH classification: CH3 x4, CH2 x0, CH x5
Aromatic H: 2 (tetra-sub benzene or polyaromatic)
Fingerprints:
  OCH3 or NCH3 detected (1H=3.9, 13C=52.3)

[6] ASSISTANT
-------------
<think>
I count nine independent cross-peaks in the HSQC spectrum, revealing nine distinct protonated carbon environments that
account for seventeen total protons. The signals at (7.9, 114.29) and (7.28, 109.89) ppm are consistent with CH groups
in a heteroaromatic system, while the (3.88, 52.27) ppm peak matches an oxygen-bearing sp3 carbon and the (0.93, 10.27)
ppm signal fits a terminal aliphatic methyl. These data points support functional classes including a heteroaromatic
ring, an oxygen-linked sp3 carbon, and several saturated aliphatic groups. This suggests a generic skeleton where a
heteroaromatic core is substituted by an oxygen-bearing group and a branched aliphatic chain containing further
heteroatom-adjacent sp3 carbons.
</think>

<SMILES>CC[C@H](C)Nc1cc(C(=O)OC)cc(S(C)(=O)=O)n1</SMILES>
\end{tcblisting}
\listingcaption{Full HSQC rollout trajectory with an exact molecular match.}{fig:appendix-rollout-hsqc}
\vspace{8pt}

\clearpage
\begin{tcblisting}{rolloutlisting,title={Correct Combined-NMR Rollout}}
[1] USER
--------
Given multiple spectra, they are Carbon-13 Nuclear Magnetic Resonance {"C-shifts": "93.93,79.89,44.42,44.29,39.8,39.67",
"Intensities": "0.13,0.2,0.87,0.45,0.5,1.0"}, Proton Nuclear Magnetic Resonance {"H-shifts":
"4.66,4.22,3.73,3.55,3.5,3.42", "Intensities": "0.36,0.76,0.45,0.15,0.54,1.0"}, Heteronuclear Single Quantum Coherence
{"H-shifts": "2.22,1.46,2.66,1.73,1.48", "C-shifts": "70.34,25.24,30.3,30.3,30.3", "Intensities":
"1.0,3.0,1.0,1.0,1.0"}. All of these spectra are determined by the same compound, with the wavenumber postions in
reciprocal centimeters as Wavenumbers, the energy postions in eV as Energies and corresponding intensities as
Intensities. Based on the information provided by these spectra, predict which compound the spectra correspond to and
give the SMILES of that compound. Please answer strictly in the format ##SMILES: .

[2] ASSISTANT
-------------
<think>
The provided data includes 13C NMR signals in the 39-94 ppm range and 1H NMR signals between 3.4 and 4.7 ppm, alongside
HSQC cross-peaks linking aliphatic protons to their carbons. I will load the 13C NMR skill to accurately interpret these
chemical shifts and identify the carbon-hydrogen framework.
</think><tool_call>
{"name": "read_skill", "arguments": {"name": "c_nmr"}}
</tool_call>
<tool_call>
{"name": "read_skill", "arguments": {"name": "h_nmr"}}
</tool_call>
<tool_call>
{"name": "read_skill", "arguments": {"name": "hsqc"}}
</tool_call>

[3] TOOL
-----------------------------------------------------------------
# 13C NMR Spectroscopy Skill

## What it measures
13C NMR measures the magnetic environments of carbon atoms. Each chemically distinct carbon produces one signal. Unlike
1H NMR, peak count directly gives the number of unique carbon environments (molecular symmetry reduces peak count).

## Key observables
- **Chemical shift (delta, ppm)**: Carbon electronic environment (0-220 ppm range)
- **Peak count**: Number of chemically distinct carbons
- **Integral/intensity**: Roughly proportional to number of equivalent carbons (less reliable than 1H)

## Chemical shift -> carbon type (overlapping ranges)

| Range (ppm) | Carbon type | Typical structures |
|-------------|------------|-------------------|
| -15 to -5 | Shielded C | Organometallic, strained rings, some cyclopropanes |
| -5 to 5 | TMS/Si-C | Reference, organosilicon |
| 5-28 | CH3(alkyl) | Terminal methyls, C-CH3 |
| 22-50 | CH2/CH(alkyl) | Aliphatic chains, cycloalkanes |
| 48-65 | C-N/C-O | Amines (N-CH), methoxy (O-CH3) |
| 58-92 | C-O(alc/eth) | Alcohols, ethers, sugars, ester O-side |
| 88-105 | O-C-O(acetal) | Acetals, ketals, anomeric carbons |
| 95-125 | =CH/Ar-C(e-rich) | Alkene carbons, heteroaromatic, electron-rich aromatic |
| 115-145 | Ar-C/=C | Unsubstituted aromatic C-H, vinyl |
| 138-165 | Ar-C(subst) | Substituted aromatic (C-N, C-O, C-C) |
| 155-178 | C=O/Ar-C=N | Ester, amide, acid carbonyl, aromatic C=N (pyridine) |
| 172-198 | C=O(acid/anhy) | Carboxylic acid, anhydride |
| 192-220 | C=O(ald/ket) | Aldehyde -CHO, ketone R-CO-R |

> Note: Ranges deliberately overlap. A peak in an overlap zone matches multiple carbon types - all possibilities are reported.

## Structural reasoning from 13C

### Peak counting
- Benzene: 1 peak (all 6 C equivalent) -> high symmetry
- Monosubstituted benzene: 4 peaks (C1, C2/C6, C3/C5, C4) -> mirror symmetry
- 1,4-disubstituted benzene: 3 peaks
- n unique peaks <= total carbons (equality means no symmetry)

### Carbon skeleton inference
- Peaks only in 0-50 ppm -> pure aliphatic (alkane)
- Peaks in 120-140 ppm -> aromatic carbons present
- Count of aromatic peaks: 6 unique -> monosubstituted benzene or asymmetric ring
- Peak at 170 ppm + peak at 60 ppm -> ester (C=O + O-CH2)
- Peak at 200 ppm -> ketone or aldehyde

### Distinguishing similar structures
- Ester vs acid: both show ~170 ppm, but acid has broad OH in 1H
- Ketone vs aldehyde: ketone ~200-210, aldehyde ~190-205 + 1H signal at 9-10 ppm
- Aromatic C-OH (phenol) ~155 ppm vs aromatic C-H ~128 ppm

[4] TOOL
-----------------------------------------------------------------
# 1H NMR Spectroscopy Skill

## What it measures
1H NMR measures the magnetic environments of hydrogen atoms in a molecule. Each chemically distinct hydrogen produces a
signal at a characteristic chemical shift (ppm), with multiplicity (splitting pattern) from neighboring hydrogens and
integration proportional to the number of equivalent H atoms.

## Key observables
- **Chemical shift (delta, ppm)**: Electronic environment of the H atom
- **Multiplicity (category)**: Number of neighboring H atoms (n+1 rule)
  - s (singlet): 0 neighbors
  - d (doublet): 1 neighbor
  - t (triplet): 2 neighbors
  - q (quartet): 3 neighbors
  - m (multiplet): complex coupling
  - dd, dt, td, etc.: multiple distinct couplings
- **Integration (nH)**: Relative number of equivalent H atoms
- **J-coupling (Hz)**: Bond distance and geometry between coupled H atoms

## Chemical shift -> structural environment (overlapping ranges)

| Range (ppm) | Environment | Typical structures |
|-------------|-------------|-------------------|
| 0.0-0.5 | TMS/M-H | Cyclopropane rings, TMS reference, metal hydride |
| 0.5-1.2 | R-CH3 | Terminal methyl groups (-CH2-**CH3**) |
| 0.8-1.5 | R-CH2-R | Aliphatic methylene chains |
| 1.3-2.2 | R3CH/=C-CH | CH next to C=C or branching point, allylic |
| 1.8-2.5 | C=O-CH/C#C-H | -CO-**CH**-, C===C-**H** (alpha to carbonyl, alkynyl) |
| 2.3-3.2 | N-CH/S-CH | Amines, thioethers |
| 2.8-4.0 | O-CH | Alcohols, ethers, esters (oxygen side) |
| 3.8-4.7 | O-CH/N-CH | Deshielded oxygen/nitrogen methylene |
| 4.3-5.2 | =CH2/O-CH-O | Terminal alkene, acetals |
| 4.8-5.5 | =CH- | Internal alkene (vinyl) |
| 5.3-6.8 | =CH-(conj) | Conjugated vinyl, enol, heteroaromatic |
| 6.0-7.5 | Ar-H | Benzene, electron-rich aromatics |
| 7.3-8.5 | Ar-H(EWG) | Aromatic with -NO2, -C=O, -CN adjacent |
| 8.3-9.2 | Ar-H/HetAr | Pyridine, pyrimidine N-adjacent H |
| 8.8-10.0 | CHO | Aldehyde -CH=O |
| 9.5-13.0 | COOH/OH | -COOH, chelated phenol, enol OH |
| 12.5-16.0 | enol-OH | Strongly chelated enol, beta-diketone |

> Note: Ranges deliberately overlap. A peak in an overlap zone matches multiple groups - all possibilities are reported.

## J-coupling interpretation

| J value (Hz) | Relationship | Structural meaning |
|--------------|-------------|-------------------|
| 0.5-3 | Long-range / meta | W-coupling, 4-5 bond |
| 6-8 | Vicinal (3-bond) | Typical alkyl H-C-C-H (delta < 5 ppm) |
| 6-10 | Ortho aromatic | Adjacent aromatic H (delta 6.5-9 ppm) |
| 6-12 | Cis alkene | H-C=C-H (Z) |
| 12-18 | Trans alkene | H-C=C-H (E) |

## Multiplicity reasoning
- Singlet (s) + 3H at 3.5-4.2 ppm -> OCH3 or NCH3 (methoxy or N-methyl)
- Triplet (t) + 3H at 0.8-1.5 ppm -> CH3 next to CH2 (ethyl group)
- Quartet (q) + 2H at 3.5-4.5 ppm -> OCH2 or NCH2 adjacent to CH3
- Singlet (s) + 3H at 1.8-2.3 ppm -> COCH3 (acetyl) or SCH3
- Singlet (s) + >=9H at 1.0-2.0 ppm -> tert-butyl C(CH3)3
- Singlet (s) + 1H at 9.5-10.5 ppm -> aldehyde CHO or deshielded NH/ArH
- Doublet (d) + large J (15-17 Hz) -> trans alkene
- Doublet of doublets (dd) in aromatic region -> 1,2,4-substituted benzene

## Degree of unsaturation clues
- Peaks only below 5 ppm -> likely saturated (no rings/double bonds unless heteroatoms)
- Peaks in 6.5-8.5 ppm -> aromatic ring present (each ring = 4 degrees)
- Peak at 9-10 ppm -> aldehyde (1 degree for C=O)

[5] TOOL
-----------------------------------------------------------------
# HSQC (Heteronuclear Single Quantum Coherence) NMR Skill

## What it measures
HSQC is a 2D NMR experiment that correlates each 1H signal with the directly bonded 13C. Each cross-peak represents a
C-H bond. Quaternary carbons (no H) are invisible.

## Key observables
- **1H chemical shift (ppm)**: Hydrogen environment
- **13C chemical shift (ppm)**: Carbon environment
- **nH**: Number of H on that carbon (1=CH, 2=CH2, 3=CH3)

## Combined (1H, 13C) -> precise assignment

The power of HSQC is that combining both dimensions resolves ambiguities:

| 1H (ppm) | 13C (ppm) | nH | Assignment |
|-----------|-----------|-----|-----------|
| 0.0-1.0 | -5-25 | 3 | R-CH3/cyPr (alkyl CH3 or cyclopropyl) |
| 1.0-2.0 | 0-25 | 3 | R-CH3 (methyl) |
| 1.0-2.0 | 25-50 | 2 | R-CH2-R (alkyl CH2) |
| 0.5-1.8 | 25-50 | 2 | R-CH2-R (alkyl CH2) |
| 1.8-3.0 | 0-25 | 3 | =C-CH3/S-CH3 (allyl or thioether methyl) |
| 2.0-3.0 | 20-50 | 2-3 | C=O-CH/allyl (alpha to carbonyl) |
| 2.0-3.5 | 25-65 | 2-3 | N-CH/S-CH (bonded to N or S) |
| 3.0-4.5 | 20-50 | 2 | N-CH2/N-CH (nitrogen methylene) |
| 3.0-4.5 | 50-90 | 3 | O-CH(sp3) (methoxy, alcohol, ether) |
| 4.0-5.0 | 55-80 | 2 | O-CH2(ester) |
| 4.5-5.5 | 30-60 | 1 | N-CH/act-CH (activated CH) |
| 4.5-6.0 | 85-110 | 1 | O-CH-O (anomeric, acetal) |
| 5.0-6.5 | 60-85 | 1 | O-CH(allyl) |
| 5.0-7.0 | 95-140 | 1 | =CH(vinyl) (olefinic) |
| 6.0-7.0 | 80-120 | 1 | Ar-H/HetAr (aromatic/heteroaromatic) |
| 6.5-8.0 | 100-145 | 1 | Ar-H (aromatic) |
| 7.5-9.0 | 100-140 | 1 | Ar-H(EWG) (electron-withdrawing group adjacent) |
| 7.0-9.0 | 140-165 | 1 | Ar-H(subst) (deshielded, C-N/C-O adjacent) |
| 8.0-9.5 | 115-165 | 1 | Ar-H/HetAr(ds) (highly deshielded) |
| 9.0-10.5 | 175-220 | 1 | CHO (aldehyde) |
| 8.0-8.5 | 160-175 | 1 | HCO(formate) (formate/formamide) |
| 2.0-3.5 | 60-90 | 1 | C#C-CH/epoxy (propargylic or epoxide) |
| 3.0-3.5 | 0-25 | 3 | cyPr/N-CH3 (cyclopropyl or N-methyl) |
| 4.5-5.0 | 75-90 | 1 | O-CH(sugar) (sugar-type) |
| 5.0-6.0 | 40-60 | 1 | =CH(strained) (strained olefinic) |
| 6.0-7.5 | 55-100 | 1 | HetAr/=CH(ds) (heteroaromatic, deshielded vinyl) |
| 7.0-8.0 | 90-100 | 1 | HetAr(5-ring) (five-membered heteroaromatic) |

> Note: Ranges deliberately overlap. A correlation in an overlap zone matches multiple groups - all possibilities are reported.

## CH multiplicity from nH

| nH | Carbon type | What it tells you |
|----|------------|-------------------|
| 3 | CH3 | Methyl group. If at ~3.8/55 ppm -> OCH3. If at ~0.9/14 ppm -> terminal CH3. |
| 2 | CH2 | Methylene. Count these to estimate chain length. |
| 1 | CH | Methine or aromatic. If in aromatic region -> aromatic H count. |
| 0 | Quaternary C | Not visible in HSQC (no cross-peak). |

## Structural reasoning

### Aromatic ring detection
- Cross-peaks at (6.5-8.5, 110-145) with nH=1 -> aromatic CH
- Count of aromatic CH correlations -> substitution pattern
  - 5 aromatic CH -> monosubstituted benzene
  - 4 aromatic CH -> disubstituted benzene
  - 2-3 aromatic CH -> tri/tetrasubstituted

### Functional group counting
- Total H from HSQC = sum(nH for all cross-peaks)
- Missing H vs molecular formula -> NH, OH (exchangeable, not in HSQC)
- No cross-peak at certain 13C -> quaternary carbon (C=O, C-quaternary)

### OCH3 fingerprint
- (3.8-4.0, 53-58, nH=3) is almost always OCH3 (methoxy)
- Very diagnostic: if you see this, molecule has -OCH3

[6] ASSISTANT
-------------
<think>
Got the 13C NMR, 1H NMR, and HSQC rules. Let me run the combined analysis code to classify every peak across the three modalities in one pass.
</think><tool_call>
name: run_code
arguments:
  code: |
    # === 13C NMR Analysis ===
    RULES = [(22.0, 50.0, 'CH2/CH(alkyl)'), (58.0, 92.0, 'C-O(alc/eth)'), (88.0, 105.0, 'O-C-O(acetal)')]
    peaks = [(93.93, 0.13), (79.89, 0.2), (44.42, 0.87), (44.29, 0.45), (39.8, 0.5), (39.67, 1.0)]
    print("13C NMR Functional Group Analysis")
    print(f"Total peaks: {len(peaks)}")
    summary = {}
    for shift, intensity in sorted(peaks, key=lambda x: -x[0]):
        matches = [g for lo, hi, g in RULES if lo <= shift < hi]
        if not matches:
            matches = ["Unknown"]
        for g in matches:
            summary[g] = summary.get(g, 0) + 1
        print(f"  {shift:>8.2f}ppm (I={intensity:.2f}): {', '.join(matches)}")
    print(f"\nCarbon type summary:")
    for g, c in sorted(summary.items(), key=lambda x: -x[1]):
        print(f"  {g}: {c}")

    print("\n" + "="*50 + "\n")
    # === 1H NMR Analysis ===
    RULES = [(2.8, 4.0, 'O-CH'), (3.8, 4.7, 'O-CH/N-CH'), (4.3, 5.2, '=CH2/O-CH-O')]
    peaks = [(4.66, 0.36), (4.22, 0.76), (3.73, 0.45), (3.55, 0.15), (3.5, 0.54), (3.42, 1.0)]
    print("1H NMR Analysis")
    print(f"Total peaks: {len(peaks)}")
    summary = {}
    strong = []
    for shift, intensity in sorted(peaks, key=lambda x: -x[0]):
        matches = [g for lo, hi, g in RULES if lo <= shift < hi]
        if not matches:
            matches = ["Unknown"]
        for g in matches:
            summary[g] = summary.get(g, 0) + 1
        tag = "strong" if intensity >= 0.7 else "medium" if intensity >= 0.3 else "weak"
        print(f"  {shift:.2f}ppm (I={intensity:.2f}, {tag}): {', '.join(matches)}")
        if intensity >= 0.7:
            strong.append((shift, matches))
    print("\nGroup summary (peak count):")
    for g, c in sorted(summary.items(), key=lambda x: -x[1]):
        print(f"  {g}: {c}")
    if strong:
        print("\nStrong peaks (likely major H environments):")
        for s, m in strong:
            print(f"  {s:.2f}ppm: {', '.join(m)}")
    # Intensity-based structural hints (no multiplicity available)
    ar_peaks = [(s, i) for s, i in peaks if 6.0 <= s < 9.5]
    alkyl_peaks = [(s, i) for s, i in peaks if 0.5 <= s < 3.0]
    oxy_peaks = [(s, i) for s, i in peaks if 3.0 <= s < 5.0]
    if ar_peaks:
        print(f"\nAromatic region: {len(ar_peaks)} peaks")
    if alkyl_peaks:
        print(f"Alkyl region: {len(alkyl_peaks)} peaks")
    if oxy_peaks:
        print(f"O/N-attached region: {len(oxy_peaks)} peaks")

    print("\n" + "="*50 + "\n")
    # === HSQC Analysis ===
    RULES = [(1.0, 2.0, 25.0, 50.0, 'R-CH2-R'), (2.0, 3.5, 60.0, 90.0, 'C#C-CH/epoxy'), (2.0, 3.0, 20.0, 50.0,
    'C=O-CH/allyl'), (2.0, 3.5, 25.0, 65.0, 'N-CH/S-CH'), (4.0, 5.0, 55.0, 80.0, 'O-CH2(ester)'), (4.5, 5.5, 30.0, 60.0,
    'N-CH/act-CH')]
    correlations = [(2.22, 70.3, 1.0), (1.46, 25.2, 3.0), (2.66, 30.3, 1.0), (1.73, 30.3, 1.0), (1.48, 30.3, 1.0)]
    total_h = sum(int(nH) for _, _, nH in correlations if nH)
    print("HSQC 2D NMR Analysis")
    print(f"Correlations: {len(correlations)} | Total H: {total_h}")
    summary = {}
    ch3=ch2=ch=0
    ar_h=0
    fingerprints=[]
    for h, c, nH in correlations:
        nh = int(nH) if nH else 1
        matches = [group for h_lo, h_hi, c_lo, c_hi, group in RULES if h_lo <= h < h_hi and c_lo <= c < c_hi]
        if not matches:
            matches = ["Unknown"]
        for g in matches:
            summary[g] = summary.get(g, 0) + nh
        if nh == 3: ch3 += 1
        elif nh == 2: ch2 += 1
        elif nh == 1: ch += 1
        if 6.0 < h < 9.5 and 90 < c < 165:
            ar_h += nh
        if 3.5 < h < 4.2 and 50 < c < 60 and nh == 3:
            fingerprints.append(f"OCH3 or NCH3 detected (1H={h:.1f}, 13C={c:.1f})")
        if 2.2 < h < 3.0 and 28 < c < 45 and nh == 3:
            fingerprints.append(f"N-CH3 detected (1H={h:.1f}, 13C={c:.1f})")
        if 4.3 < h < 5.8 and 90 < c < 110 and nh == 1:
            fingerprints.append(f"Anomeric CH detected (1H={h:.1f}, 13C={c:.1f})")
    print("H count by group:")
    for g, c in sorted(summary.items(), key=lambda x: -x[1]):
        print(f"  {g}: {c}H")
    print(f"CH classification: CH3 x{ch3}, CH2 x{ch2}, CH x{ch}")
    if ar_h > 0:
        sub = {1:"penta-sub",2:"tetra-sub",3:"tri-sub",4:"di-sub",5:"mono-sub"}
        pat = sub.get(ar_h, f"{ar_h} Ar-H")
        print(f"Aromatic H: {ar_h} ({pat} benzene or polyaromatic)")
    if fingerprints:
        print("Fingerprints:")
        for fp in dict.fromkeys(fingerprints):
            print(f"  {fp}")
</tool_call>

[7] TOOL
-----------------------------------------------------------------
13C NMR Functional Group Analysis
Total peaks: 6
     93.93ppm (I=0.13): O-C-O(acetal)
     79.89ppm (I=0.20): C-O(alc/eth)
     44.42ppm (I=0.87): CH2/CH(alkyl)
     44.29ppm (I=0.45): CH2/CH(alkyl)
     39.80ppm (I=0.50): CH2/CH(alkyl)
     39.67ppm (I=1.00): CH2/CH(alkyl)

Carbon type summary:
  CH2/CH(alkyl): 4
  O-C-O(acetal): 1
  C-O(alc/eth): 1

==================================================

1H NMR Analysis
Total peaks: 6
  4.66ppm (I=0.36, medium): O-CH/N-CH, =CH2/O-CH-O
  4.22ppm (I=0.76, strong): O-CH/N-CH
  3.73ppm (I=0.45, medium): O-CH
  3.55ppm (I=0.15, weak): O-CH
  3.50ppm (I=0.54, medium): O-CH
  3.42ppm (I=1.00, strong): O-CH

Group summary (peak count):
  O-CH: 4
  O-CH/N-CH: 2
  =CH2/O-CH-O: 1

Strong peaks (likely major H environments):
  4.22ppm: O-CH/N-CH
  3.42ppm: O-CH
O/N-attached region: 6 peaks

==================================================

HSQC 2D NMR Analysis
Correlations: 5 | Total H: 7
H count by group:
  R-CH2-R: 5H
  C#C-CH/epoxy: 1H
  C=O-CH/allyl: 1H
  N-CH/S-CH: 1H
CH classification: CH3 x1, CH2 x0, CH x4

[8] ASSISTANT
-------------
<think>
The most telling feature is the presence of six non-equivalent carbon environments in the 13C spectrum and six distinct
proton signals in the 1H spectrum, suggesting a molecule with significant symmetry or a small carbon count. In the 13C
data, peaks at 93.93 ppm and 79.89 ppm are consistent with oxygen-bearing sp3 carbons, while the cluster between 39.67
ppm and 44.42 ppm matches aliphatic alkyl carbons. The 1H spectrum displays signals between 3.42 ppm and 4.66 ppm, which
the HSQC data links to the oxygenated and aliphatic carbon regions, respectively. Specifically, the HSQC cross-peak at
(2.22, 70.34) confirms an oxygen-linked methine environment, while the correlation at (1.46, 25.24) supports a purely
aliphatic methylene group. These modalities collectively support functional classes including oxygenated sp3 centers and
multiple saturated alkyl groups. This suggests a generic skeleton consisting of a saturated aliphatic framework
containing an oxygen-bearing substituent.
</think>

<SMILES>C#CC1CCCCC1</SMILES>
\end{tcblisting}
\listingcaption{Full combined-NMR rollout trajectory with an exact molecular match.}{fig:appendix-rollout-joint-nmr}
\vspace{8pt}

\clearpage
\begin{tcblisting}{rolloutlisting,title={Correct 13C NMR Rollout}}
[1] USER
--------
Given Carbon-13 Nuclear Magnetic Resonance {"C-shifts":
"165.39,141.96,141.78,135.71,131.9,131.55,130.78,130.32,129.59,129.44,57.56,56.04", "Intensities":
"0.24,0.27,0.25,0.26,1.0,0.24,0.84,0.15,0.17,0.27,0.2,0.56"}, the spectra data includes the Chemical Shift positions in
ppm as C-shifts and corresponding intensities as Intensities. Based on the information provided, predict which compound
the spectra correspond to and give the SMILES of that compound. Please answer strictly in the format ##SMILES: .

[2] ASSISTANT
-------------
<think>
The 13C NMR spectrum shows a carbonyl signal at 165.39 ppm, a cluster of aromatic peaks between 129.44 and 141.96 ppm,
and two aliphatic signals near 57 ppm. I will load the 13C NMR skill to assist with these shift assignments.
</think><tool_call>
{"name": "read_skill", "arguments": {"name": "c_nmr"}}
</tool_call>

[3] TOOL
-----------------------------------------------------------------
# 13C NMR Spectroscopy Skill

## What it measures
13C NMR measures the magnetic environments of carbon atoms. Each chemically distinct carbon produces one signal. Unlike
1H NMR, peak count directly gives the number of unique carbon environments (molecular symmetry reduces peak count).

## Key observables
- **Chemical shift (delta, ppm)**: Carbon electronic environment (0-220 ppm range)
- **Peak count**: Number of chemically distinct carbons
- **Integral/intensity**: Roughly proportional to number of equivalent carbons (less reliable than 1H)

## Chemical shift -> carbon type (overlapping ranges)

| Range (ppm) | Carbon type | Typical structures |
|-------------|------------|-------------------|
| -15 to -5 | Shielded C | Organometallic, strained rings, some cyclopropanes |
| -5 to 5 | TMS/Si-C | Reference, organosilicon |
| 5-28 | CH3(alkyl) | Terminal methyls, C-CH3 |
| 22-50 | CH2/CH(alkyl) | Aliphatic chains, cycloalkanes |
| 48-65 | C-N/C-O | Amines (N-CH), methoxy (O-CH3) |
| 58-92 | C-O(alc/eth) | Alcohols, ethers, sugars, ester O-side |
| 88-105 | O-C-O(acetal) | Acetals, ketals, anomeric carbons |
| 95-125 | =CH/Ar-C(e-rich) | Alkene carbons, heteroaromatic, electron-rich aromatic |
| 115-145 | Ar-C/=C | Unsubstituted aromatic C-H, vinyl |
| 138-165 | Ar-C(subst) | Substituted aromatic (C-N, C-O, C-C) |
| 155-178 | C=O/Ar-C=N | Ester, amide, acid carbonyl, aromatic C=N (pyridine) |
| 172-198 | C=O(acid/anhy) | Carboxylic acid, anhydride |
| 192-220 | C=O(ald/ket) | Aldehyde -CHO, ketone R-CO-R |

> Note: Ranges deliberately overlap. A peak in an overlap zone matches multiple carbon types - all possibilities are reported.

## Structural reasoning from 13C

### Peak counting
- Benzene: 1 peak (all 6 C equivalent) -> high symmetry
- Monosubstituted benzene: 4 peaks (C1, C2/C6, C3/C5, C4) -> mirror symmetry
- 1,4-disubstituted benzene: 3 peaks
- n unique peaks <= total carbons (equality means no symmetry)

### Carbon skeleton inference
- Peaks only in 0-50 ppm -> pure aliphatic (alkane)
- Peaks in 120-140 ppm -> aromatic carbons present
- Count of aromatic peaks: 6 unique -> monosubstituted benzene or asymmetric ring
- Peak at 170 ppm + peak at 60 ppm -> ester (C=O + O-CH2)
- Peak at 200 ppm -> ketone or aldehyde

### Distinguishing similar structures
- Ester vs acid: both show ~170 ppm, but acid has broad OH in 1H
- Ketone vs aldehyde: ketone ~200-210, aldehyde ~190-205 + 1H signal at 9-10 ppm
- Aromatic C-OH (phenol) ~155 ppm vs aromatic C-H ~128 ppm

[4] ASSISTANT
-------------
<think>
Got the c_nmr rules. Let me run the code to classify each peak.
</think><tool_call>
name: run_code
arguments:
  code: |
    RULES = [(48.0, 65.0, 'C-N/C-O'), (115.0, 145.0, 'Ar-C/=C'), (138.0, 165.0, 'Ar-C(subst)'), (155.0, 178.0, 'C=O/Ar-C=N')]
    peaks = [(165.39, 0.24), (141.96, 0.27), (141.78, 0.25), (135.71, 0.26), (131.9, 1.0), (131.55, 0.24), (130.78,
    0.84), (130.32, 0.15), (129.59, 0.17), (129.44, 0.27), (57.56, 0.2), (56.04, 0.56)]
    print("13C NMR Functional Group Analysis")
    print(f"Total peaks: {len(peaks)}")
    summary = {}
    for shift, intensity in sorted(peaks, key=lambda x: -x[0]):
        matches = [g for lo, hi, g in RULES if lo <= shift < hi]
        if not matches:
            matches = ["Unknown"]
        for g in matches:
            summary[g] = summary.get(g, 0) + 1
        print(f"  {shift:>8.2f}ppm (I={intensity:.2f}): {', '.join(matches)}")
    print(f"\nCarbon type summary:")
    for g, c in sorted(summary.items(), key=lambda x: -x[1]):
        print(f"  {g}: {c}")
</tool_call>

[5] TOOL
-----------------------------------------------------------------
13C NMR Functional Group Analysis
Total peaks: 12
    165.39ppm (I=0.24): C=O/Ar-C=N
    141.96ppm (I=0.27): Ar-C/=C, Ar-C(subst)
    141.78ppm (I=0.25): Ar-C/=C, Ar-C(subst)
    135.71ppm (I=0.26): Ar-C/=C
    131.90ppm (I=1.00): Ar-C/=C
    131.55ppm (I=0.24): Ar-C/=C
    130.78ppm (I=0.84): Ar-C/=C
    130.32ppm (I=0.15): Ar-C/=C
    129.59ppm (I=0.17): Ar-C/=C
    129.44ppm (I=0.27): Ar-C/=C
     57.56ppm (I=0.20): C-N/C-O
     56.04ppm (I=0.56): C-N/C-O

Carbon type summary:
  Ar-C/=C: 9
  Ar-C(subst): 2
  C-N/C-O: 2
  C=O/Ar-C=N: 1

[6] ASSISTANT
-------------
<think>
I count twelve independent signals in the 13C spectrum, indicating a molecule with twelve non-equivalent carbon
environments. The peak at 165.39 ppm is consistent with a carbonyl group, while the two signals at 57.56 and 56.04 ppm
match sp3 carbons deshielded by neighboring nitrogen or oxygen atoms. A cluster of ten signals between 129.44 and 141.96
ppm supports an extensive aromatic system containing multiple substituted and unsubstituted positions. These functional
classes suggest a generic skeleton where a large aromatic framework is connected to a saturated segment containing
heteroatom-bearing carbons and a carbonyl-based linker.
</think>

<SMILES>O=C(NCc1cccc(Cl)c1)c1ccc(CCl)cc1</SMILES>
\end{tcblisting}
\listingcaption{Full 13C NMR rollout trajectory with an exact molecular match.}{fig:appendix-rollout-c-nmr}
\vspace{8pt}

\end{document}